\documentclass{article}
\usepackage[top=1in, bottom=1in, left=1in, right=1in]{geometry}
\usepackage[utf8]{inputenc}

\usepackage{url}

\usepackage{microtype}
\usepackage{graphicx}
\usepackage{booktabs} %
\usepackage{hyperref}
\usepackage{overpic}

\usepackage{caption}
\usepackage[numbers,sort&compress]{natbib}

\usepackage{amsmath,amssymb,amsfonts,amsthm}

\usepackage{mathrsfs}
\usepackage{xcolor}
\usepackage{mathtools}
\usepackage{dsfont}
\usepackage{hyperref}
\usepackage{bm}
\usepackage{nicefrac}
\usepackage{wrapfig}
\usepackage{lipsum}
\usepackage{enumerate}

\usepackage{algorithm, algcompatible}
\usepackage{algpseudocode}
\algnewcommand\algorithmicinput{\textbf{Input:}}
\algnewcommand\INPUT{\item[\algorithmicinput]}
\algnewcommand\algorithmicoutput{\textbf{Output:}}
\algnewcommand\OUTPUT{\item[\algorithmicoutput]}
\algnewcommand\algorithmicoptional{\textbf{Optional:}}
\algnewcommand\OPTIONAL{\item[\algorithmicoptional]}
\let\oldReturn\Return
\renewcommand{\Return}{\State\oldReturn}

\newcounter{assumption}%
\renewcommand{\theassumption}{\arabic{assumption}}

\newcommand*\E[1]{\mathbb{E}\left[#1\right]}
\newcommand*\Ep[2]{\mathbb{E}_{#1}\left[#2\right]}

\newcommand*\lrbb[1]{\left\{#1\right\}}
\newcommand*\lrp[1]{\left(#1\right)}
\newcommand*\lrn[1]{\left\|#1\right\|}

\newcommand*\ind[1]{{\mathds{1}\lrbb{#1}}}

\newcommand{\real}{\ensuremath{\mathbb{R}}}

\newcommand{\ve}[1]{\mathbf{#1}}

\newtheorem{lemma}{Lemma}
\newtheorem{theorem}{Theorem}

\newtheorem{remark}{Remark}

\numberwithin{lemma}{section} 
\numberwithin{theorem}{section} 
\numberwithin{corollary}{section} 
\numberwithin{proposition}{section} 
\numberwithin{definition}{section} 
\numberwithin{example}{section} 
\numberwithin{question}{section} 

\usepackage{multirow}
\usepackage{caption}
\usepackage{subcaption}

\newtheoremstyle{bfnoteonly}%
{}{}%
{\itshape}{}%
{\bfseries}{.}%
{ }%
{\thmnote{#3}}

\theoremstyle{bfnoteonly}

\newcommand{\Xv}{\mathbf{X}}

\newcommand{\fv}{\mathbf{f}}

\newcommand{\xv}{\mathbf{x}}

\newcommand{\thetav}{\mathbf{\theta}}
\newcommand{\Thetav}{\boldsymbol{\Theta}}

\newcommand{\Xiv}{\boldsymbol{\Xi}}
\newcommand{\xiv}{\boldsymbol{\xi}}

\newcommand\Rb{\mathbb{R}}

\title{Convergence of uncertainty estimates in \\ Ensemble and Bayesian sparse model discovery}

\author{L. Mars Gao$^*$, Urban Fasel$^{**}$, Steven L. Brunton$^\dag$ and J. Nathan Kutz$^\ddag$\\[.2in]
$^*$Paul G. Allen School of Computer Science \& Engineering, 
University of Washington\\
$^{**}$Department of Aeronautics, Imperial College, London SW7 2AZ, United Kingdom\\
$^\dag$Department of Mechanical Engineering, University of Washington, Seattle, WA 98195\\
$^\ddag$Department of Applied Mathematics and Electrical and Computer Engineering,\\ University of Washington, Seattle, WA 98195}
\date{\today}

\begin{document}

\maketitle

\begin{abstract}
    Sparse model identification enables nonlinear dynamical system discovery from data. 
    However, the control of false discoveries for sparse model identification is challenging, especially in the low-data and high-noise limit. 
    In this paper, we perform a theoretical study on ensemble sparse model discovery, which shows empirical success in terms of accuracy and robustness to noise.   
    In particular, we analyse the bootstrapping-based sequential thresholding least-squares estimator. 
    We show that this bootstrapping-based ensembling technique can perform a provably correct variable selection procedure with an exponential convergence rate of the error rate. 
    In addition, we show that the ensemble sparse model discovery method can perform computationally efficient uncertainty estimation, compared to expensive Bayesian uncertainty quantification methods via MCMC. 
    We demonstrate the convergence properties and connection to uncertainty quantification in various numerical studies on synthetic sparse linear regression and sparse model discovery. The experiments on sparse linear regression support that the bootstrapping-based sequential thresholding least-squares method has better performance for sparse variable selection compared to LASSO, thresholding least-squares, and bootstrapping-based LASSO.
    In the sparse model discovery experiment, we show that the bootstrapping-based sequential thresholding least-squares method can provide valid uncertainty quantification, converging to a delta measure centered around the true value with increased sample sizes. Finally, we highlight the improved robustness to hyperparameter selection under shifting noise and sparsity levels of the bootstrapping-based sequential thresholding least-squares method compared to other sparse regression methods. 
\end{abstract}

\section{Introduction}

Data-driven model discovery methods based on sparse linear regression have been demonstrably successful, enabling the discovery of interpretable and generalizable models that balance accuracy and efficiency. The sparse identification of nonlinear dynamics (SINDy)~\citep{brunton2016discovering} model discovery framework can leverage a variety of sparsity promoting algorithms, including the {\em sequential thresholding least-squares} (STLS) estimator introduced in the original paper.  Recently, SINDy was extended to incorporate ensembling techniques, leveraging bootstrap aggregating (bagging) methods~\cite{breiman1996bagging} to produce the robust and computationally efficient probabilistic model discovery method  {\em ensemble SINDy} (E-SINDy)~\citep{fasel2022ensemble}. With the bootstrapping step, E-SINDy empirically improves the robustness of variable selection when limited and noisy data is available, and enables uncertainty quantification in parametric inference. 
Here, we consider sparse linear regression using bagging inclusion probability for STLS estimation, a generalization of the sparse regression algorithm used in E-SINDy. We prove that the E-SINDy algorithm is able to perform correct sparse model identification with exponential convergence in terms of false discovery probability (FDP) and true discovery probability (TDP). The theoretical result on the correct sparse identification additionally offers a statistical guarantee on the uncertainty quantification of E-SINDy. These theoretical results reveal the effectiveness of E-SINDy for sparse model discovery, and its potential as a general sparse linear regression method.


Variable selection and sparse regression have a rich history in data science and statistics. In classical theory, the likelihood ratio test is one of the earliest variable selection methods~\citep{wilks1938large}. More recently, the least absolute shrinkage and selection operator (LASSO) uses an $\ell_1$ shrinkage estimator for sparse inference~\citep{tibshirani1996regression}. 
The statistical properties of LASSO-based methods have been systematically studied, including the asymptotic behavior~\citep{fu2000asymptotics}, consistency~\citep{zhao2006model}, the oracle property in classic regression settings~\citep{fan2001variable,zou2006adaptive}, and in high-dimensional settings~\citep{wainwright2009sharp,wainwright2019high,huang2008asymptotic}. 
From the Bayesian perspective, a hierarchical Bayesian model with spike-and-slab~\citep{mitchell1988bayesian,george1997approaches,ishwaran2005spike}, Laplace~\citep{park2008bayesian}, or regularized horseshoe prior~\citep{carvalho2009handling,carvalho2010horseshoe} have been thoroughly studied for Bayesian sparse inference. 
For thresholding-based estimators, even if these estimators have been widely applied in problems like wavelet regression~\citep{donoho1994ideal}, denoising~\citep{donoho1995noising}, robust high-dimensional regression~\citep{bhatia2015robust}, and sparse identification of physical systems~\citep{brunton2016discovering}, the development of their rigorous statistical properties have not been fully established. The first work on statistical guarantees for STLS based estimators investigated the oracle property of STLS estimators in low- and high-dimensional settings, and showed the consistency of residual-bootstrap for STLS estimators~\citep{giurcanu2016thresholding}. 
Characterizing the effect of bootstrapping applied to sparse variable selection adds another layer of complexity. 
In general, bootstrapping is a computational method to improve inference on small samples using the idea of resampling~\citep{efron1979computers,hall2013bootstrap,efron1994introduction}. The residual-bootstrap LASSO is inconsistent for sparse models~\citep{chatterjee2011bootstrapping}, but the residual-bootstrap distribution of Adaptive LASSO~\citep{zou2006adaptive} is consistent. Resampling methods can help improve variable selection performance for LASSO~\citep{bach2008bolasso,meinshausen2010stability,werner2022loss}, but there has only been limited studies that jointly consider bootstrapping and STLS. To the best of our knowledge, the detailed interaction between bootstrap and STLS still remains unknown.

\begin{figure*}[t]
    \includegraphics[width=\textwidth]{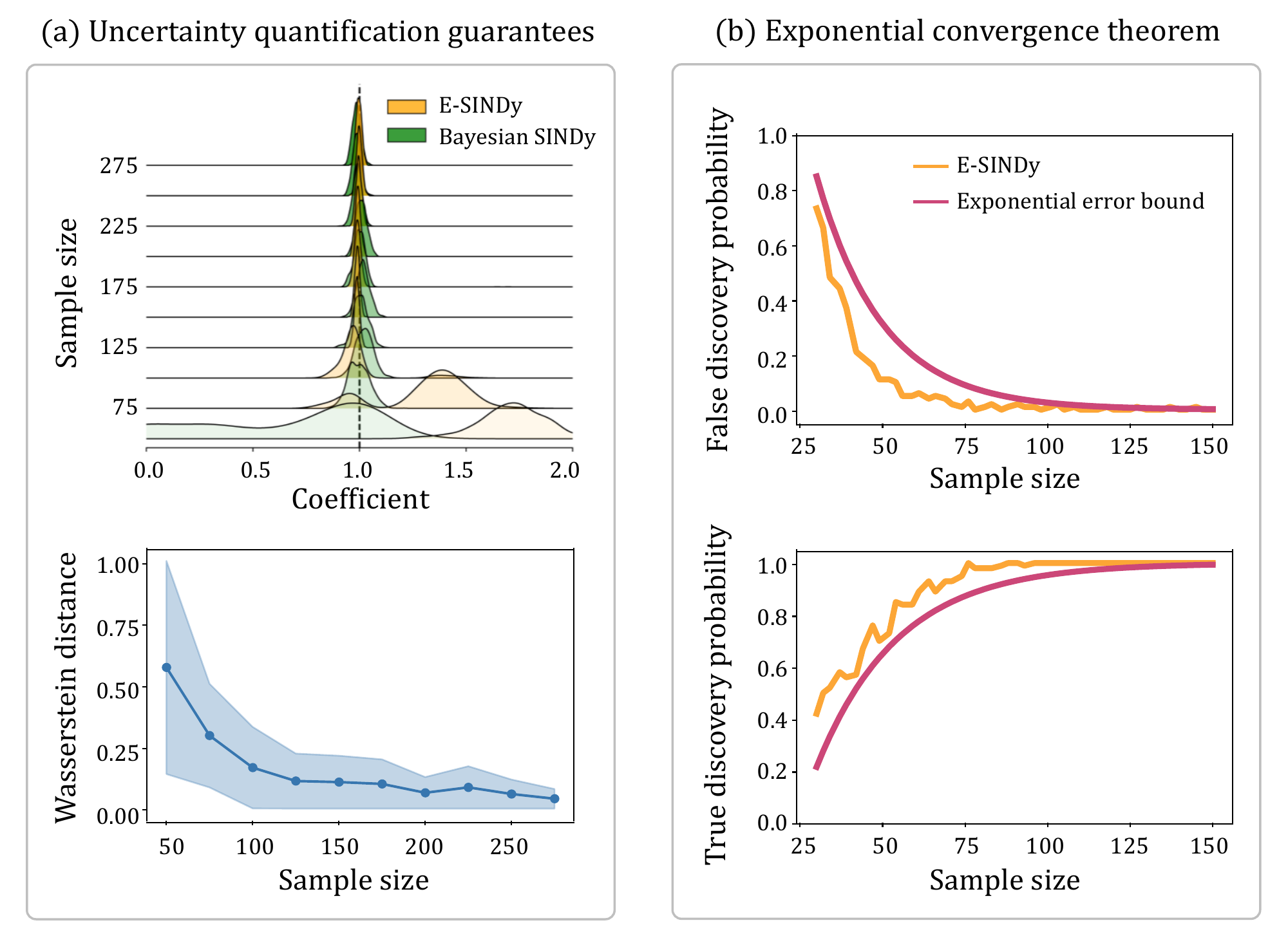}
    \caption{(a) Top: convergence of distributional estimation of the Bayesian SINDy (green) and E-SINDy algorithm (orange) over different sample sizes. Bottom: Wasserstein-2 distance between E-SINDy and Bayesian SINDy with Spike-and-slab prior. 
    (b) Exponential convergence for E-SINDy in terms of False Discovery Probability and True Discovery Probability for the synthetic Lotka-Volterra dataset. 
    }
    \label{fig:conv_and_uq}
\end{figure*}


To establish the theoretical understanding of the interaction between bootstrapping and STLS, we investigate the bootstrap aggregating inclusion probability based STLS estimation. 
Through concentration probability bounds, we show the oracle property in Thm.~\ref{thm:oracle_property} of bootstrapping inclusion probability STLS with an exponential convergence rate under milder regularity conditions. 
This theoretical result guarantees that the bootstrapping inclusion probability STLS guarantees the identification of the correct model when having sufficiently large sample sizes. 
In Fig.~\ref{fig:conv_and_uq} (b), the exponential convergence of error bounds and the empirical convergence using E-SINDy for the FDP and TDP are shown. 

In addition to the oracle property for variable selection, we show that the E-SINDy algorithm with the bagging inclusion probability based STLS estimator is valid for uncertainty quantification (UQ) in Theorem~\ref{lem:uq_BIP}. We show the similarity between bagging inclusion probability STLS and Bayesian UQ methods with spike-and-slab prior~\citep{ishwaran2005spike,hirsh2022sparsifying} from (i) asymptotic convergence in Wasserstein distance~\citep{newton1994approximate,ishwaran2005spike} and (ii) Bayesian bootstrap Spike-and-Slab LASSO~\citep{nie2022bayesian}. 
Compared to Bayesian UQ methods, bootstrapping based methods can provide similar uncertainty estimates, but are computationally much more efficient. For example, in the Lotka-Volterra model discovery~\cite{hirsh2022sparsifying}, Bayesian SINDy requires a computational execution time of more than 12 hours, while E-SINDy takes less than 20 seconds. We correspondingly discuss the difference in computational complexity between Bayesian MCMC and bagging inclusion probability-based methods in Remark~\ref{rem:computation}. 
The effectiveness of E-SINDy for distributional convergence and uncertainty quantification is shown in Fig.~\ref{fig:conv_and_uq} (a) top. The E-SINDy algorithm quickly converges to the true mean with an increasing number of samples, and generates an uncertainty estimate around the true value. In Fig.~\ref{fig:conv_and_uq} (a) bottom, the blue solid line represents the average Wasserstein-2 distance over ten experimental trials. The shaded area shows the confidence interval estimate using the $80\%$ percentile over the ten trials. The distributional estimation of E-SINDy converges to the estimation of Bayesian SINDy with large sample sizes. This behavior of convergence is examined under Wasserstein distance in Thm.~\ref{thm:w2_convergence} from the asymptotic sense.  



We further apply bagging inclusion probability based STLS to broader scenarios, from synthetic sparse linear regression to physics model discovery. From the simulation of sparse linear regression under different settings, we observe an accelerated effect of bagging inclusion probability based variable selection in terms of the sample size. Additionally, we perform an experimental study on the robustness of hyperparameters in varying experimental settings for bagging inclusion probability based STLS and other baseline methods. The bagging inclusion probability based STLS has very stable performances with respect to changes in sparsity and noise levels. 
For governing equation discovery, we observe that the E-SINDy based algorithm can accurately perform variable selection with uncertainty quantification. The estimation bias quickly reduces with more observation points and the uncertainty estimation is valid under noisy settings. The experimental study, as the supplement of Thm.~\ref{lem:uq_BIP}, shows that the E-SINDy with bagging inclusion probability based STLS uncertainty quantification is valid, which is functionally equivalent to Bayesian MCMC with sparsifying priors~\citep{hirsh2022sparsifying}. 

The contributions of this paper are threefold: 

\begin{enumerate}
    \item We study the theoretical extension of E-SINDy algorithm with bagging inclusion probability STLS and analyze its oracle property under regularity conditions with an exponential convergence rate. 
    \item We show that bagging inclusion probability STLS is asymptotically equivalent to the Bayesian method with lower computational cost, and can perform valid uncertainty quantification with statistical guarantee. 
    \item We implement and verify the effect of bagging inclusion probability STLS in both synthetic sparse linear regression settings and physical equation discovery. We verify the effect of accelerated variable selection as well as the validity in uncertainty quantification. 
\end{enumerate}

In what follows, we first introduce the SINDy algorithm and variable selection methods in Sec.~\ref{sec:background}. We then review the analysis on sequential thresholding least-squares estimation in Sec.~\ref{sec:tlse}, and propose a theoretical analysis for E-SINDy with bootstrapping inclusion probability based STLS in Sec.~\ref{sec:bs_tlse}. We utilize Sec.~\ref{sec:uq_bootstrapvalidity} to provide the uncertainty quantification guarantees and connections to the Bayesian method. Finally,  we conduct various experiments on sparse linear regression and sparse physics model discovery in Sec.~\ref{sec:simulation}. 
\section{Background}
\label{sec:background}

\subsection{Sparse identification of nonlinear dynamics}

The {\em sparse identification of nonlinear dynamics} (SINDy) \cite{brunton2016discovering} algorithm is a thresholding least-square sparse regression technique to identify the underlying dynamical system from observational data snapshots $\xv(t) \in \Rb^n$ where
\begin{align}
  \dot{\ve{x}}(t) = \fv(\xv(t)).
    \label{eq:dynamical_system_x}
\end{align}
The data snapshots $\xv(t)$ are measured at time $t$, and the function $\fv$ represents the underlying dynamical system that is to be discovered. By calculating the temporal derivatives from the snapshot data, SINDy forms the following covariates and targets:
\begin{align}
  \Xv = \left(\begin{array}{cccc}
    \xv (t_1) \\
    \xv (t_2) \\
    \vdots \\
    \xv (t_m) \\
  \end{array}\right), \quad
  \dot{\Xv} = \left(\begin{array}{cccc}
    \dot{\xv}_1(t_1) \\
    \dot{\xv}_1(t_2) \\
    \vdots \\
    \dot{\xv}_1(t_m) \\
  \end{array}\right). \nonumber
\end{align}
A function library is constructed with $p$ candidate model terms so that $\Thetav(\Xv) = [\thetav_1(\Xv) \cdots \thetav_p(\Xv)] \in \Rb^{m\times p}$. 
The candidate functions can be polynomials, sinusoids, exponentials, and so on. 
In summary, we hope to derive a sparse model between $\ve{X}$ and $\dot{\ve{X}}$ such that
\begin{equation}
    \label{eq:sindy_definition}
  \dot{\Xv} = \Thetav(\Xv)\Xiv
\end{equation}
where the unknown sparse matrix $\Xiv = (\xiv_1\ \xiv_2\ \cdots\ \xiv_n)\in \Rb^{p\times n}$ enables the correct identification of the dynamics. 
Typically, SINDy~\cite{brunton2016discovering} applies STLS estimation to perform sparse inference, which is a proxy for $\ell_0$ optimization~\cite{zheng2018unified} with convergence guarantees~\cite{zhang2019convergence}.  
Hirsh et al.~\citep{hirsh2022sparsifying} perform Bayesian sparse inference in model discovery via Spike-and-slab and regularized horseshoe priors.  
Fasel et al.~\citep{fasel2022ensemble} combine bootstrap resampling to perform uncertainty estimation with observed acceleration in variable selection for system identification.  
SINDy has been applied to diverse physical systems 
including fluid dynamics~\cite{loiseau2018constrained,loiseau2018sparse,loiseau2020data,guan2021sparse,deng2021galerkin,callaham2022empirical}, nonlinear optics~\cite{sorokina2016sparse}, turbulence
closures~\citep{beetham2020formulating,beetham2021sparse,schmelzer2020discovery}, ocean closures~\citep{zanna2020data}, chemical reactions~\cite{hoffmann2019reactive},  plasma dynamics~\cite{dam2017sparse,alves2022data,kaptanoglu2021physics}, structural modeling~\cite{lai2019sparse}, granular materials~\cite{zhao2022data}, and for model predictive control~\cite{kaiser2018sparse}. 
Extensions of SINDy includes the identification of partial differential equations~\cite{rudy2017data,schaeffer2017learning}, parametrically dependent dynamical models~\cite{rudy2019data},  multiscale physics~\cite{champion2019discovery},  time-dependent PDEs~\citep{chen2021robust}, rational function nonlinearities~\cite{mangan2016inferring,kaheman2020sindy}, switching dynamical systems~\cite{mangan2019model}, control inputs~\citep{kaiser2018sparse}, constraints on symmetries~\citep{loiseau2018constrained}, control for stability~\citep{kaptanoglu2021promoting}, control for robustness~\citep{alves2022data,schaeffer2017sparse,reinbold2020using,reinbold2021robust,messenger2021weak,messenger2021weak}, stochastic dynamical systems~\citep{boninsegna2018sparse,callaham2021nonlinear}, multidimensional approximation on tensors~\citep{gelss2019multidimensional}, and video data in pixel space~\citep{champion2019data,bakarji2022discovering,gao2022bayesian}.  

\subsection{Variable selection, bootstrap resampling, and uncertainty quantification}

From the perspective of uncertainty quantification,  sparse models have different approaches for uncertainty estimation. The bootstrap method is one way to obtain an uncertainty estimate of sparse regression. Chatterjee and Lahiri~\citep{chatterjee2010asymptotic} first showed residual-bootstrap of LASSO is inconsistent. Chatterjee and Lahiri~\citep{chatterjee2011bootstrapping} then showed that the residual-bootstrap of adaptive LASSO estimator is asymptotically consistent.
Extending from the bootstrap approximation, Bayesian methods are more favorable for obtaining uncertainty estimates of sparse models with sparsifying priors. 
Mitchell and Beauchamp~\citep{mitchell1988bayesian} first established Spike-and-slab models.  These were further developed by George and McCulloch~\citep{george1997approaches}. Ishwaran and Rao proposed the general theoretical framework of Spike-and-slab prior in~\citep{ishwaran2005spike}. 
Carvalho, Polson, and Scott utilized the horseshoe prior~\citep{carvalho2009handling,carvalho2010horseshoe} to perform Bayesian sparse inference. 
Computations of the above-mentioned Bayesian methods rely heavily on  hierarchical MCMC sampling.  
Ročková and George~\citep{rovckova2018spike} proposed Spike-and-slab LASSO that focuses on mode searching instead of traversing the entire distribution. Nie and Ročková~\citep{nie2022bayesian} applied Bayesian bootstrap to Spike-and-slab LASSO to obtain uncertainty estimation with a first-order correctness guarantee. 
The Spike-and-slab LASSO method is a thresholding procedure~\citep{nie2022bayesian} (c.f. Eqn. (4)). 
Giurcanu established the analysis to show the consistency of residual-bootstrap for TLSE~\citep{giurcanu2016thresholding}. 

Bootstrap resampling and aggregation can frequently improve the performance of unstable estimators. Bühlmann and Yu~\citep{buhlmann2002analyzing} examined this idea through case studies in bagging with indicators and subsampling aggregating decision trees. 
Bühlmann and Yu~\citep{buhlmann2006sparse} also introduced sparse boosting to obtain good performance in $\ell_2$. 
Later in Bolasso~\citep{bach2008bolasso}, Bach proposed a model consistent LASSO method via bootstrapping with improved performances in variable selection. 
Meinshausen and Bühlmann established stability-based selection which provides a consistent variable selection process~\citep{meinshausen2010stability}. They showed the error control bound of false discovery variables and theoretical guarantee of randomized LASSO algorithm under sparse eigenvalue condition. Recently, Werner~\citep{werner2022loss} proposed loss-guided stability selection. 


\section{Sequential thresholding least-square in sparse regression}
\label{sec:tlse}

In this section, we establish the theoretical foundations of the STLS estimator in a sparse linear model. We constraint our analysis to the primary effect of STLS by only performing the thresholding procedure twice from the OLS estimate, which is more accessible for theoretical analysis. 

Consider a dataset $\{\ve{X}_i, \ve{y}_i\}_{i=1}^n$ that is described by a linear model
\begin{align}
    \ve{y}_i = \ve{X_i}^T \ve{\beta} + \ve{\epsilon}_i,
\end{align}
where $\ve{y}_i\in\real$ is the response; $\ve{X}_i^T \in\real^p$ is the covariate; $\beta\in\real^p$ is an unknown parameter; and $\ve{\epsilon}_i\overset{\mathrm{iid}}{\sim} P$ is the noise following a distribution $P$ on $\real$ satisfying $\E{\epsilon}=0$, and $Var(\epsilon)=\sigma^2<\infty$. For simplicity, we center the response and standardize covariates so that $\bar{Y}=0, \bar{X}^{(j)}=0$, and $\bar{S}^{(j)}=1$. 
In the context of SINDy (Eqn.~\eqref{eq:sindy_definition}), the function library $\mathbf{\Theta(X)}$ is the covariate $\ve{X}_i^T$, $\mathbf{\dot{X}}$ is the response $\ve{y}_i$, and $\mathbf{\Xi}$ is the unknown parameter $\beta$.

For system identification, we wish to obtain a sparse model that is the most explainable. We set $q$ to be the number of non-zero components of $\beta$ so that $q<p$. In our setting, we consider the model size $p$ has similar scale with the sample $n$ under sparsity constraints on $\beta$. 
We focus on the case when $n>p$ with independent experimental trials.

\subsection{Thresholding least-squares estimate}
Let $\bar\beta$ be the ordinary least-squares (OLS) solution of $\beta$ such that
\begin{align}
    \bar\beta = (\ve{X}^T\ve{X})^{-1}\ve{X}^T\ve{y}.
\end{align}
Let $\rho_1, \rho_2$ be the minimum and the maximum eigenvalues of $\frac{1}{n}\ve{X}^T\ve{X}$. Given the regularity conditions in~\eqref{lem:assumption}, one can achieve asymptotic normality via Lindeberg-Feller CLT without tail assumptions.

\begin{lemma}
\label{lem:ols_tail}
Given the regularity conditions in the Appendix~\eqref{lem:assumption}, then $\forall a\in\real^p$, 

\begin{align}
    \label{lemeq:normality}
    a^T(\bar{\beta}-\beta)\xrightarrow{d}\mathcal{N}\lrp{0,\sigma^{2}a^T(\ve{X}^T\ve{X})^{-1}a},
\end{align}
and for $\delta\in (0,1)$ we have
\begin{align}
    \label{lemeq:tail_bound}
    \mathbb{P}\lrp{a^T(\bar\beta-\beta) \geq \sqrt{2\sigma^{2}\log(1/\delta)a^T(\ve{X}^T\ve{X})^{-1}a}} \leq \delta.
\end{align}
\end{lemma}

We show a proof of Lemma~\ref{lem:ols_tail} in the Appendix. 
From Lemma~\ref{lem:ols_tail}, we can have a tail bound of the OLS with probability at least $1-\delta$, 
\begin{align}
    a^T(\bar\beta-\beta) \leq \sigma\sqrt{2\log(1/\delta)a^T(\ve{X}^T\ve{X})a}.
\end{align}
However, without further assumption on the noise, the tail behavior of the OLS is only valid in the asymptotic sense. In this section, to demonstrate the rate of convergence in the non-asymptotic sense, we make a stronger assumption that emphasizes the role of Gaussian noise with a known variance. 

\vspace{3mm}
(A.1) $\epsilon\sim\mathcal{N}(0, \sigma^2I)$. 
\vspace{3mm}



Let $\hat{K}$ be the thresholding estimator of the non-active index set of zero components $K_\beta$. Motivated by the Gaussian tail bound, we set 
\begin{align}
    \hat K = \{j\in I: |\bar\beta_j| \leq \sqrt{2\sigma^2\log(1/\delta)(\ve{X}_j^T\ve{X}_j)^{-1}}\}. 
\end{align}
The sequential thresholding least-squares estimator in our paper ($\hat\beta$) is defined as the following: 
\begin{align}
    \label{def:tlse}
    \hat\beta = \arg\min_{b\in\real^p} \left\{(\ve{y}-\ve{X}b)^T(\ve{y}-\ve{X}b):b_{\hat{K}}=0\right\}.
\end{align}
Equivalently, we can write $\hat\beta$ as
\begin{align}
    \begin{cases}
    \hat\beta_{\hat{K}}=0,\\
    \hat\beta_{\hat{J}} = \lrp{\ve{X}^T_{\hat{J}}\ve{X}_{\hat{J}}}^{-1}\ve{X}^T_{\hat{J}}\ve{y},
    \end{cases}
\end{align}
where $\hat{J} = I \backslash \hat{K}$. 

From \cite{giurcanu2016thresholding} (c.f. Theorem 2.1), STLS has the oracle property. We present the following Lemma under (A.1) in the following. 
\begin{lemma}
\label{lem:tlse_oracle}
Suppose that A.1, (a), (b) hold, and $n\rho_1^2\to\infty$. Then $\hat\beta$ has the oracle property $Pr(\hat K = K_\beta)=1$ since the following holds: 

(i) The False Discovery Probability (FDP) satisfies 
    \begin{align}
        1-Pr\lrp{K_\beta \subset \hat K_{STLS}}\leq\frac{q}{n}. 
    \end{align}

(ii) The True Discovery Probability (TDP) satisfies 
    \begin{align}
        Pr\lrp{\hat K_{STLS} \subset K_\beta} \geq 1- 2
    \exp\lrp{-\frac{1}{2}\lrp{\frac{n^{1/2}\rho_1^{1/2}|\beta_{(1)}|}{\sigma}- \sqrt{2\log(n)}-\sqrt{2\log(p-q_{K_\beta})}}^2}.
    \end{align}

\end{lemma}

\section{Bagging sequential thresholding least-square estimator}
\label{sec:bs_tlse}

Bootstrapping is a computational technique to perform uncertainty quantification, variance reduction, and bias estimation. 
The benefit of  the bootstrap aggregating estimator~\cite{breiman1996bagging} often comes from unstable statistics \citep{buhlmann2002analyzing} (c.f. Definition 1.2). In the case of the STLS estimator, the inclusion of an indicator function makes  the STLS an unstable statistics. As stated in Sec. 2 in \citep{buhlmann2002analyzing}, we expect to see a boosted effect from Bagging due to the smoothing instead of the hard thresholding operation. We present the Bagging inclusion probability based STLS estimate along with the estimation via inclusion probability which can also be understood as a selection probability \citep{meinshausen2010stability}. We analyze and claim that the inclusion probability based variable selection can achieve the same oracle property with less restrictive assumptions on sample size. 

\subsection{Bagging inclusion probability STLS}

\begin{algorithm*}[t]
    \caption{{Bagging inclusion probability thresholding least-square estimation}}
    \label{alg:bag_tlse}
    \begin{algorithmic}[1]
    \INPUT{covariate $X$, target $y$, thresholding constant $\gamma$, cv proportion $c$, inclusion probability threshold $p$.}
    \OUTPUT{a BIP estimate of active subset $\tilde J$}
    \Function{BaggingInclusionProbability}{$X, y$}
        \State $IncludeList=List()$;   \Comment{create an empty list}
        \For{i in $0, 1, \cdots, n-1$:}
            \State $X_{B}, y_{B} = Subsample(X, y, c)$;
            \State $\beta_{B} = LeastSquares(X_{B}, y_{B})$; \Comment{compute least-square estimate given train data}
            \State $threshold = \sigma\sqrt{\frac{\gamma}{\text{diag}(X_{B}^T X_{B})}}$;  \Comment{threshold array for $\beta_{BS}$ from train data}
            \State $include = (\beta_{BS} > threshold)$;  \Comment{decide whether to include a variable or not from threshold}
            \State $IncludeList.append(include)$;
        \EndFor
        \State $InclusionProb = \sum_{i=1}^n IncludeList[i]$;
        \State $\hat{J} = InclusionProb > p$; \Comment{select indices based on inclusion probability threshold $p$}
        \Return $\hat{J}$
    \EndFunction
    \end{algorithmic}
\end{algorithm*}

%
%
%

%
%


Following the standard bootstrap procedure, we resample $n$ bootstrap replicates and estimate $\hat K_{STLS}$, denoted by $\{\hat K^b_{STLS}\}_{b=1}^n$. We apply the standard bootstrap procedure to facilitate later analysis of uncertainty quantification in Sec.~\ref{sec:uq_bootstrapvalidity}. This setting alternatively considers the bagging inclusion probability of $j\in I$ with a constant thresholding probability $p_c$ 
where
\begin{align}
    \label{eqn:BIP}
    \hat K_{BIP} = \left\{j\in I: \frac{1}{n}\sum_{b=1}^n \ind{|\bar\beta^b_j| > \sigma\sqrt{2\log(1/\delta)(\ve{X}_j^{bT}\ve{X}^b_j)^{-1}}} \leq p_c\right\}. 
\end{align}


The idea of bagging inclusion probability estimation $\hat K_{BIP}$ (demonstrated in Algorithm~\ref{alg:bag_tlse}) is very straightforward. We count the number of index $j$ that appear in all bootstrap replicates $\hat K^b_{TLS}$ and divide the count by $n$. If the inclusion probability for an index $j$ is low (e.g. $p_j<0.05$), we say $j$ is not active since most bootstrap replicates do not contain $j$. On the other hand, we say the index $j$ is active when the inclusion probability is high. 

\subsection{The analysis on bagging inclusion probability estimation}
In the following, we show that bagging Inclusion Probability estimator $\hat K_{BIP}$ has the oracle property that, with a sufficient number of samples, $\hat K_{BIP}$ will converge to the correct $K_\beta$. The regularity condition for the number of samples is given under the following assumptions:

(B.1) Let $q$ be the number of non-zero components and $p$ be the model size. We assume $q,p$ are of the same scale and that the number of samples are much larger than the model size so that
\begin{align}
    \frac{q}{p}=O(1), \;\;\;\;\frac{p}{\exp(n)} = o(1). 
\end{align}
The condition (B.1) is necessary for our main Theorem. This assumption on sample complexity is much more relaxed comparing to \citep{giurcanu2016thresholding}. 

(B.2) We assume $n$ is relatively large which satisfies 
\begin{align}
    \frac{\sqrt{n}\lrp{|\beta_{1}|-\max_{j\in J_\beta}|\bar\beta_j-\beta_j|}}{\sigma\sqrt{(r_0\rho_1)^{-1}}} \geq \sqrt{2\log(n)},
\end{align}
where $|\beta_1| = \min_{j\in J_\beta} |\beta_j|$, and $r_0$ is a constant that $r_0\rho_1$ is smaller than the minimal eigenvalue of all bootstrap replicates $X^b$. 

Condition (B.2) depicts the case when we have a large sample size and $\sqrt{2\log(1/n)(\ve{X}_j^{bT}\ve{X}_j^b)^{-1}}$ is relatively small comparing to the magnitude of $\beta$. Therefore, under (C.2), a non-sparse component is more likely to be correctly identified as active. 

(B.3) We also consider sub-Gaussian tails \citep{zhang2020non}. Suppose a centered random variable $X$ is finite for a range of $t\in T\subseteq \real$. There exist constants $c_1,C_1>0$ such that
\begin{align}
    \exp\lrp{c_1(\E{X}^2)t^2} \leq \E{\exp(tX)} \leq \exp\lrp{C_1(\E{X}^2)t^2},
\end{align}
holds for $t$ in a neighbourhood of 0. 
The assumption (B.3) helps to depict the behavior when (B.2) is unachievable. This consideration can help us understand how $\hat K_{BIP}$ performs with very small sample size. This assumption (B.3) is motivated from Taylor's expansion $\E{\exp(tX)} = 1+t^2\E{X}^2+o(t^2)$. The upper bound makes sense as being similar to the form of a sub-gaussian tail bound. The lower bound is also reasonable to assume for tail behaviors like Gaussian. 

We define the {\em False Discovery Probability} (FDP) as follows. Given the non-active set estimation as $\hat K$, the FDP is defined as
\begin{align}
    \label{def:FDP}
    \text{FDP} = 1-Pr\lrp{K_\beta \subset \hat K}.
\end{align}
The FDP characterizes the probability of whether the current estimate $\hat K$ contains false discovery. 
This metric is stronger compared to the {\em False discovery rate} (FDR) due to the fact that the FDR must be 0 when $K_\beta\subset\hat K$ is true. 

Similarly, we define the {\em True Discovery Probability} (TDP) such that 
\begin{align}
    \label{def:TDP}
    \text{TDP} = Pr\lrp{\hat K \subset K_\beta }.
\end{align}
The TDP characterizes the probability of whether the current estimate $\hat K$ contains all true active coefficients. This metric is also stronger compared to the {\em true positive proportion} (TPP) due to the fact that the TPP must be 1 when $\hat K\subset K_\beta$ is true. 

We use the following theorem to show the goodness of the bootstrap inclusion probability estimator $\hat K_{BIP}$. 

\begin{theorem}
\label{thm:oracle_property}
Given assumption A.1, we have the following: 
\begin{enumerate}[(i)]
    \item The False Discovery Probability (FDP) satisfies 
    \begin{align}
        1-Pr\lrp{K_\beta \subset \hat K_{BIP}}\leq (p-q)\exp\lrp{\frac{1}{3}-\frac{np_c}{3}} = \mathcal{O}\lrp{\frac{p}{\exp(n)}}. 
    \end{align}
    \item Under large sample size condition (B.2), the True Discovery Probability (TDP) satisfies 
    \begin{align}
        Pr\lrp{\hat K_{BIP} \subset K_\beta} \geq 1- q\exp\lrp{-\frac{n(1-2p_c)^2}{6}} = 1-\mathcal{O}\lrp{\frac{q}{\exp(n)}}.
    \end{align}
    \item For small sample size condition, under (B.3), the True Discovery Probability (TDP) satisfies 
    \begin{align}
        Pr\lrp{\hat K_{BIP} \subset K_\beta} \geq 1- q\exp\lrp{-\frac{np_c^2}{3p}+\frac{2np_c}{3}-\frac{np}{3}} =1- \mathcal{O}\lrp{\frac{q}{\exp(n)}}.
    \end{align}
    \item Under (B.1), we have the Oracle Property that 
    \begin{align}
        Pr\lrp{\hat K_{BIP} = K_\beta} \to 1. 
    \end{align}
\end{enumerate}
\end{theorem}

Details of the proof are left to the Appendix. 
We achieve this proof largely by the Gaussian tail assumption associated with the Hoeffding's bound. Generally, one could expect the least-square estimate of sparse coefficients will shrink towards 0. The inclusion probability estimated by the 
resampling-based replication will tend to be more accurate as the sample size goes up, so the sparse subset estimation will become more robust given a reasonable threshold probability. 

An important result from Thm.~\ref{thm:oracle_property} is that $\hat K_{BIP}$ is a very good estimator of $K_\beta$ with an exponential convergence rate for both FDP and TDP.
In the proof, we implicitly assume the existence of a threshold probability $p_c$. We validate the general existence of $p_c$ in the following Lem.~\ref{lem:threshold} with further details in the discussion. Compared to STLS, given similar conditions, the rate of convergence can only reach exponential behavior on one side (either FDP or TDP) but not both. The exponential rate shows rapid decay in FPD and rapid increase in TPD in Sec. 6. Utilizing this good estimation on $K_\beta$, one expects to achieve accurate inference on the unknown parameter $\beta$ with many estimators.

\paragraph{Determine a threshold probability.} 
It is not trivial to determine the hyperparameter settings for sparse regression methods. 
For example, LASSO depends heavily on the $\ell_1$ constraint and STLS requires a reasonable least-square threshold. 
BIP provides an alternative to determine the threshold probability via data analysis. The inclusion probability usually shows a clear gap between active and non-active indices. We define this gap as $\Delta$. We can formalize this idea by extending Thm.~\ref{thm:oracle_property} in the following Lemma. 
\begin{lemma}
\label{lem:threshold}
Given the assumption and conditions in Theorem~\ref{thm:oracle_property}, the inclusion probability gap $\Delta$ satisfies $\Delta > \epsilon$ with probability at least $1-pe^{-n\lrp{\frac{1}{4}-\frac{\epsilon}{2}-\frac{1}{2n}}^2/2}$ where $\epsilon$ is a constant that $\epsilon\in (0,0.5)$. 
\end{lemma}

The previous Lemma~\ref{lem:threshold} is derived via a similar process to the proof of  Thm.~\ref{thm:oracle_property}. Consider the case when $p=30, n=100$, according to the bound, we have with probability at least $97\%$ that we can observe the inclusion probability gap between the minimum of active indices and the maximum of non-active indices greater than $0.2$. By running the algorithm once, one could clearly determine the active set of indices, and utilize this information to select the threshold probability. We further wish to note that in practice, the gap should be even more obvious compared to the above Lemma - as we simplified the discussion of $\mathcal{I}_1$ in the proof. In numerical experiments, the gap could be very significant where we observe the inclusion probabilities of active indices are empirically lower bounded by $0.99$, and non-active indices empirically are upper bounded by $0.01$. Even under cases with exceptionally high observation noise, it is accessible to perform cross-validation and use the threshold that has the lowest validation error.


\section{Uncertainty quantification with bootstrap}
\label{sec:uq_bootstrapvalidity}
In practice, it is essential to understand the uncertainty of parameter inference. Bootstrapping is one of the most classical method to access the uncertainty via computational methods. Bootstrap estimate of uncertainty are frequently preferred for "non-asymptotic" sample size~\citep{wakefield2013bayesian}. We first recall the residual-bootstrap inference process from~\citep{freedman1981bootstrapping,efron1994introduction}. Define sample residuals $ E_{1:n}=\{\epsilon_1, ..., \epsilon_n\}$ that $\epsilon_i = \ve{y}_i-\ve{X}_i^T\hat\beta$. Let $\mathbb{P}=n^{-1}\sum_{i=1}^n \delta_{\epsilon_i}$ to be the empirical distribution where $\delta_{\epsilon_i}$ is a point mass. Here, $E_{1:n}$ is the empirical distribution of $E$. By sampling with replacement, we obtain a bootstrap sample $E_{1:m}^*=\{\epsilon_1^*, ..., \epsilon_m^*\}$ which consists of $m$ conditional i.i.d. samples from $\mathbb P$. 
Then we apply the resampled residuals to the target via $\ve{y}_i^*=\ve{X}_i^T\hat\beta+\epsilon_i^*$. 

A very natural way is to construct bootstrap replicates using $\arg\min_{b\in\real^p} \left\{\lrp{\ve{y}^*-\ve{X}b}^T(\ve{y}^*-\ve{X}b):b_{\hat K^*}=0\right\}$ where $\hat{K}^*$ is estimated from different bootstrap replicates. 
However, in this case, each bootstrap replicate will potentially have different sparse subset estimation $\hat{K}^*$, which could be troublesome to analyze in theory.
Therefore, we utilize $\hat{K}_{BIP}$ defined in the previous section which could similarly make use of all the bootstrap replicates and, at the same time, avoid the randomized behavior of bootstrapped subset estimation in theory. 
In other words, we consider the estimator $\hat\beta^*$ is computed via  that 
\begin{align}
    \label{eqn:bootstrap}
    \hat\beta^* = \arg\min_{b\in\real^p} \left\{\lrp{\ve{y}^*-\ve{X}b}^T(\ve{y}^*-\ve{X}b):b_{\hat K_{BIP}}=0\right\}
\end{align}

From each bootstrap replicate $\hat\beta^*$, the starred residuals are defined as $\hat\epsilon_{1:m}^* = \ve{y}^*-\ve{X}\hat\beta^*$. Notice it is different from the setting in prior work~\citep{giurcanu2016thresholding} which only utilizes $\hat\beta$. The bootstrap version of standard error estimates are 
\begin{align*}
    {\mu}_{m}^* = \frac{1}{m} \sum_{i=1}^m \hat\epsilon_{i}^* \;\;\;\;\text{and}\;\;\;\;\hat{\sigma}^{*^2}_{m}=\frac{1}{m} \sum_{i=1}^m \hat\epsilon_{i}^{*^2}-\mu_m^{*^2}.
\end{align*}

From the bootstrap estimate of the standard error, we define $\hat{s}_{m}$ as
\begin{align}
    \hat{s}_{m} = n^{-1} \hat{\sigma}_{m}^{*^2} a_{\hat{J}}^T\lrp{X_{\hat{J}}^TX_{\hat{J}}}^{-1}a_{\hat{J}}.
\end{align}


\subsection{Consistency of bootstrap inference}

To understand the consistency of bootstrap inference in an asymptotic sense. It facilitates an asymptotically correct distributional estimation of the pivot quantity $\sqrt{n}\lrp{\hat\beta-\beta}$ from Monte Carlo samples of $\sqrt{n}\lrp{\hat{\beta}^*-\hat\beta}$. Based on the selection probability, one could propose bootstrap replicates using $\hat\beta^*$ based on $\hat{K}_{BIP}$. 
%
The bagging inclusion probability based bootstrapping uncertainty quantification does not require any extra distributional assumptions, which is an important advantage for practical applications. 
Under the following conditions, we can see the bootstrap distribution $\mathcal{L}(\hat{s}_{m}^{*-1/2}a^T(\hat\beta_m^*-\tilde\beta)\mid \ve{y})$ is consistent. 

\vspace{3mm}
(C.1) $\E{\epsilon}=0$, $Var(\epsilon)=\sigma^2<\infty$. 

\vspace{3mm}

\begin{theorem}
\label{lem:uq_BIP}
Suppose (C.1) (D.2) and (c) (d) (in the Appendix) hold. We have
\begin{align}
    \mathcal{L}(\hat{s}_{m}^{-1/2}a^T(\hat\beta^*_m-\hat\beta)|\ve{y}) \xrightarrow[]{Pr} \mathcal{N}(0, 1), 
\end{align}
for $\beta\in\Theta$. 
\end{theorem}

We demonstrate the proof of Thm.~\ref{lem:uq_BIP} in the Appendix~\ref{app:proof_uq_BIP} motivated from~\citep{giurcanu2016thresholding} by utilizing the Lindeberg-Feller Central Limit Theorem. 
With the selection from the bagging inclusion probability, compared to prior works, we achieve this proof without any extra distributional assumption on $\epsilon$.
The asymptotic consistency provides theoretical evidence to employ bagging inclusion probability for uncertainty quantification. 

\subsection{Connection to Bayesian uncertainty quantification}
\label{sec:connection_bayesian}

Bayesian MCMC with sparsifying priors gives a full Bayesian approach for sparse model inference~\citep{ishwaran2005spike,hirsh2022sparsifying}. Exemplified with the Spike-and-slab prior, we use a hierarchical prior such that
\begin{align}
    \beta_j \mid \lambda_j &\sim \lambda_j \mathcal{N}(0, c^2) + (1-\lambda_j) \mathcal{N}(0, \epsilon^2), \nonumber\\
    \lambda_j &\sim Ber(\pi),
\end{align}
with normal (lognormal) likelihood model (c.f. Eqn. (3.3) and (3.4) in ~\citep{hirsh2022sparsifying}). A classical result shows that the unrescaled Spike and slab model admits the form of generalized ridge regression, and its posterior mean asymptotically converges in probability to the OLS. This implies that the spike and slab model estimation also converges in distribution to the OLS estimate. Therefore, we have the following Theorem. 

\begin{theorem}
\label{thm:w2_convergence}
Suppose (C.1) (D.2) and (c) (d) in the Appendix hold, the asymptotic distribution of bagging inclusion probability estimation $\hat\beta^*$ and spike and slab model estimate $\tilde\beta_{SSL}$ satisfies 
\begin{align}
    \mathcal{W}_2^2(\mathcal{L}(\hat\beta^*), \mathcal{L}(\tilde\beta_{SSL})) \to 0.
\end{align}
\end{theorem}

Here, $\mathcal{W}_2^2(\cdot,\cdot)$ denotes the Wasserstein-2 distance which is a metric to measure the distance between two measures in the probability space. The convergence in Wasserstein distance demonstrates that the Bayesian spike and slab distributional estimate is asymptotically equivalent to the bagging inclusion probability distributional estimate.

Another viewpoint is from the combination of Bayesian bootstrap Spike-and-Slab LASSO~\citep{rovckova2018spike} using Expectation-Maximization Variable Selection~\citep{rovckova2014emvs}.  
The Spike-and-Slab LASSO is a thresholding procedure. The threshold of index $j$ is estimated by $\Delta_j=\inf_{t>0}\lrp{||X_j||^2t/2 - \sigma^2 \lrp{-\lambda_1|t|+\log\lrp{\frac{p^\star_\theta(0)}{p^\star_\theta(t)}}}}$, where $p^\star_\theta(t)=\frac{\theta\psi_1(t)}{\theta\psi_1(t)+(1-\theta)\psi_0(t)}$,  $\psi_0(\cdot)$ denotes a spike distribution, and $\psi_1(\cdot)$ denotes a slab distribution. 
The TLS estimate approximates $\Delta_j$ by $\gamma\bar{\sigma}_{jj}$ which can be understood as a carefully constructed $\psi_0(\cdot), \psi_1(\cdot)$. The bootstrapping procedure is an approximation to the Bayesian bootstrap~\citep{rubin1981bayesian}. Therefore, the bootstrapping STLS is an approximation of a Bayesian bootstrap Spike-and-Slab LASSO process (BB-SSL). The BB-SSL can perform uncertainty quantification with a minimax-optimal concentration rate (c.f. Thm. 4.1~\citep{nie2022bayesian}), which can also serve as a connection to the distributional estimate of the Bayesian. 

A major drawback of the Bayesian Spike-and-slab prior comes from its computational requirement. We formulate the following Remark to state the computational difference. 

\begin{remark}
\label{rem:computation}
Bayesian MCMC with Spike-and-slab model has computational complexity for each iteration that scales as $\Omega(n^2p)$. Thus Bagging Inclusion Probability has computational complexity for each bootstrap replicate of $\mathcal{O}\lrp{np^2}$. 
\end{remark}


\begin{table}[t]
  \centering
  \begin{tabular}{lcccccc}
    \toprule
       & $n=50$ & $n=100$ & $n=150$ & $n=200$ & $n=250$ \\
    \midrule
    Bayesian & $2.9\times 10^4$ & $4.1\times 10^4$ & $5.8\times 10^4$ & $4.9\times 10^4$ & $3.4\times 10^4$  \\
    Ensemble & $12.7$ & $17.0$ & $17.3$ & $17.8$ & $18.0$\\
    \bottomrule
  \end{tabular}
  \vspace{-.5em}
  \caption{The computational difference in CPU time (seconds) for Bayesian SINDy and Ensemble SINDy.} \label{tab:compute_time}
\end{table}

The computational complexity of bootstrap is much lower compared to Bayesian MCMC with Spike-and-slab prior. To obtain $n$ posterior samples from the Spike-and-slab model, the lower bound of computational complexity scales with $\Omega(n^3)$. Alternatively, the bootstrap approximation only requires a computational upper bound $\mathcal{O}(n^2)$. In real experiments for physical model discovery, as shown in Table,~\ref{tab:compute_time}, the expected running time of the spike-and-slab model can be much higher than $n^3$ due to the unpredictable rejection rate from the MCMC. Typically, performing Bayesian spike-and-slab SINDy with $150$ samples would require around $16$ hours, which is over $3,000$ times more expensive compared to the execution time of Ensemble SINDy (around $17$ seconds). 

\section{Experiments}
\label{sec:simulation}

In the following subsections, we apply bagging inclusion probability STLS to various cases, including synthetic linear regression and physical model discovery. 
We first investigate sparse linear regression under different data generation models. Compared to baseline methods, the bagging inclusion probability STLS outperforms other methods in terms of variable selection. 
Additionally, we study the robustness of hyperparameters in changing noise and sparsity levels. Bagging inclusion probability STLS has very stable performances with respect to the changing experimental settings while other methods fail to achieve stable performance. 
For governing equation discovery, we observe E-SINDy with bagging inclusion probability STLS can perform correct variable selection and uncertainty quantification. The estimation bias quickly reduces with more observation points and the uncertainty estimation is valid under noisy settings. 
\subsection{Implementation}

In the previous discussion, we considered the bagging inclusion probability selection in theory. In practice, there are many variants of bootstrap Thresholding Least-squares that fit better to real data. 
One strategy is to construct common statistics from the bootstrap samples, like the mean, median, and maximum. For example, we could consider the mean and variance of bootstrap samples to perform hypothesis testing. Also, we may utilize the maximum of the absolute value of $\hat\beta^*$ with a threshold and select variables based on the inclusion probability. Constructing different statistics may have different advantages and drawbacks from case to case. 

In the numerical simulation, we specifically utilize a re-weighted bagging inclusion probability selection using the out-of-bag (OOB) error which empirically achieves the best results. Instead of naively putting each bootstrap sample with equal weights, re-weighting each sample based on its testing error helps to improve the performance of variable selection. 
There have been similar ideas in loss-guided stability selection~\citep{werner2022loss} and cross-validation variable selection~\citep{shao1993linear}. 
Intuitively, the mean squared error of a bootstrap replicate estimate should be low when the subset selection is perfect but can be very high when the subset is incorrect.
Therefore, by adding the information from the testing loss, the bootstrap replicates can be utilized in a better way. This variant enjoys a similar theoretical guarantee as Thm.~\ref{thm:oracle_property}. 
From our experimental study, re-weighting with OOB error can help to improve the performance experimentally. We demonstrate the implementation in Algorithm~\ref{alg:bag_tlse_oob} which can be found in Appendix~\ref{app:oob_implementation}.

\subsection{Simulation of linear models: consistent variable selection}
\subsubsection{Simulation model setup}
\label{sec:simulation_model_setup}
To compare fairly to the previous models, we first examine our model based on standard model generation as in~\citep{giurcanu2016thresholding,huang2008asymptotic}. 
Then, we incorporate more challenging settings with very noisy environments. 
We also report a False discovery rate study following the setup from~\citep{su2017false}. 
The implementation is based on Python with NumPy for STLS, bootstrap STLS, and sklearn for LASSO. 


We consider the following models from $\ve{X}\in\real^{n\times p}$ with
\begin{align}
\label{eqn:syn_data_gen}
    \ve{y} = \ve{X}^T\beta + \epsilon,
\end{align}
where $\epsilon\sim\mathcal{N}(0, \sigma)$. 

\paragraph{Model 1. [Linear regression, independent correlation structure]} We set the data to $p=30$, $\sigma=1.0$, $X_i\sim\text{i.i.d. } \mathcal{N}(0, I_{30})$.
We set the coefficient $\beta$ to $\beta_{1:15}=0.0$, $\beta_{16:30}=1.0$. 

\paragraph{Model 2. [Linear regression, mild toeplitz correlation structure]} We set the data to $p=30$, $\sigma=0.6$, $X_i\sim\text{i.i.d. } \mathcal{N}(0, \Sigma)$ where $\Sigma_{ij} = r^{|i-j|}$, $r=0.3$. 
We set the coefficient $\beta$ to $\beta_{1:15}=0.0$, $\beta_{16:20}=0.5$, $\beta_{21:25}=1.5$, $\beta_{25:30}=2.5$. 

\paragraph{Model 3. [Sparse regression for model discovery]} We generate the design matrix $X$ using the following process. First generate $z=[z_1, z_2]$ where $z\sim\mathcal{N}(0, I_{2\times 2})$. From $z$, construct $X=\Theta(z)$ such that 
\[
\Theta(z)=[1, z_1, z_2, z_1^2, z_1z_2, z_2^2, z_1^3, z_1z_2^2, z_1^2z_2, z_2^3, z_1^4, z_1^3z_2, z_1^2z_2^2, z_1z_2^3,z_2^4]^T. 
\]
Finally, we use $\beta\in\real^{15\times 2}$ to generate $y\in\real^{n\times 2}$ following \eqref{eqn:syn_data_gen}. We study a synthetic Lotka-Volterra setting with $\beta[1,0]=1.0$, $\beta[5,0]=-0.68$, $\beta[1,1]=-1.5$, $\beta[5,1]=0.82$ and all other terms set to zero. Note that for this setting, we directly generate the temporal derivative target using general sparse regression settings. 





\paragraph{Evaluation metrics} We study the three simulation models from a basic linear regression, mild correlation structure~\citep{huang2008asymptotic}, and sparse symbolic regression. 
We access the performance of variable selection by the defined metric that follows. 





We consider a classical metric in variable selection using the empirical relative frequency of correct identification. For index $j$, the relative frequency of correct identification is defined as
\begin{align}
    \hat{p}_j = \begin{cases}
    S^{-1} \sum_{s=1}^S \ind{\hat\beta_j^s=0}, \; j\in K_\beta \\
    S^{-1} \sum_{s=1}^S \ind{\hat\beta_j^s \neq 0}, \; j\notin K_\beta
    \end{cases},
\end{align}
where $S$ is the total number of simulations, $\hat\beta^s=\lrp{\hat\beta^s_1, \dots, \hat\beta^S_P}^T$ is the estimation of $s$-th simulation. 


\subsubsection{Simulation outcomes}
\begin{figure}
    \raggedleft
    \begin{overpic}[width=0.95\textwidth]{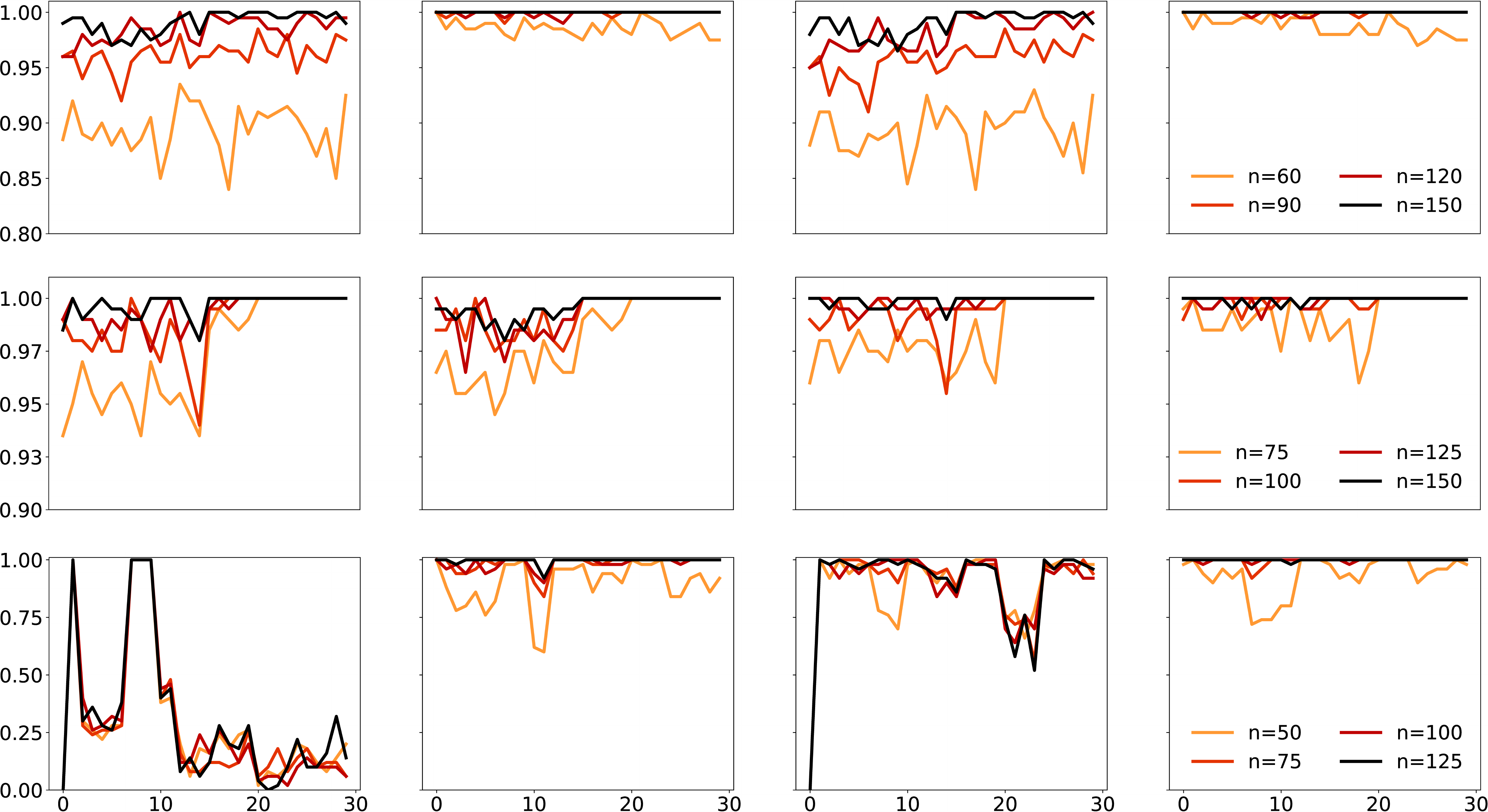}
    \put(10,55.0){\normalsize LASSO}
    \put(36,55.0){\normalsize STLS}
    \put(60,55.0){\normalsize BLASSO}
    \put(85.5,55.0){\normalsize BSTLS}
    \put(12,-2.0){\normalsize Index}
    \put(37,-2.0){\normalsize Index}
    \put(62,-2.0){\normalsize Index}
    \put(87,-2.0){\normalsize Index}
    \put(-2.5,5.5){\rotatebox{90}{\normalsize Model 3}}
    \put(-2.5,23.5){\rotatebox{90}{\normalsize Model 2}}
    \put(-2.5,42.5){\rotatebox{90}{\normalsize Model 1}}
    \put(-6,16.5){\rotatebox{90}{\large Relative frequency}}
    \end{overpic}
    \vspace{1.0mm}
    \caption{Relative frequency of three synthetic data generation models. Subfigures in the first line show the simulation outcome of Model 1; subfigures in the second line show the simulation outcome of Model 2; subfigures in the third line show the simulation outcome of Model 3. }
    \label{fig:fig:rel_freq}
\end{figure}
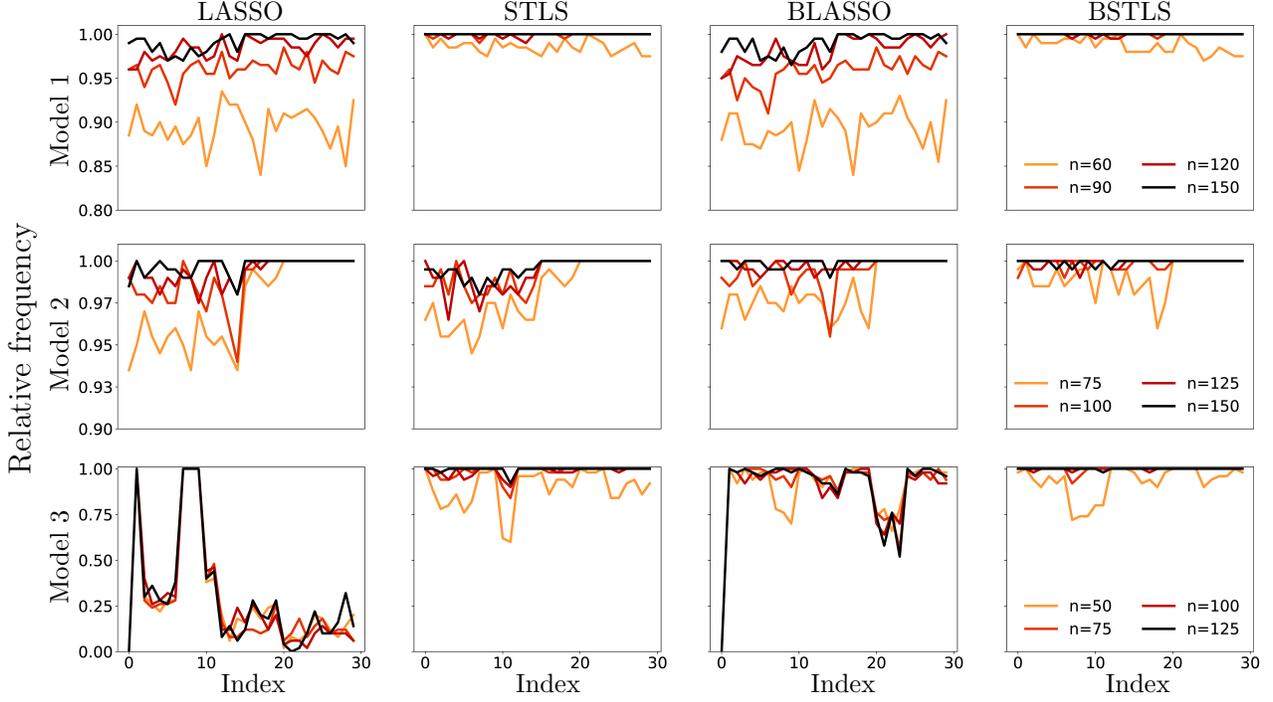

In Fig.~\ref{fig:fig:rel_freq}, we show the simulation outcome of the three simulation models. All models are fine-tuned in the following process. We first select the optimal hyper-parameter of LASSO and STLS. Then, we perform the bootstrapping inclusion probability selection process based on the fine-tuned LASSO and STLS. Finally, we decide the inclusion probability threshold separately for bagging inclusion probability based LASSO and bagging inclusion probability based STLS. We replicate the experiments $200$ times to compute the relative frequency. 
\begin{itemize}
    \item Model 1 generates independent experimental trails with $1.0$ signal-to-noise ratio. We set $\gamma=0.15p\log(p)$ for STLS, $\lambda=0.4$ as $\ell_1$ constraint in LASSO. For both bagging inclusion probability based LASSO and bagging inclusion probability based STLS, we subsample $80\%$ of data and set the inclusion probability threshold to be $0.45$. 
    Comparing LASSO and STLS, STLS has better performance with smaller sample size, but is relatively worse when the sample size is large. For Bagging inclusion probability selection, both bagging inclusion probability based LASSO and bagging inclusion probability based STLS improve the performance of LASSO and STLS respectively. bagging inclusion probability based STLS outperforms all other methods for all sample sizes. 
    \item Model 2 generates design with mild correlation structure. We set $\gamma=0.1p\log(p)$ for STLS, $\lambda=0.2$ as $\ell_1$ constraint in LASSO. For both bagging inclusion probability based LASSO and bagging inclusion probability based STLS, we subsample $80\%$ of data and set the inclusion probability threshold to be $0.7$. 
    STLS has better performances with increasing number of samples compared to LASSO. When $n=60$, notice the minimum of empirical relative frequency of STLS is $98\%$ while LASSO could only reach around $85\%$. The sample complexity of LASSO in this case is clearly higher compared to STLS since LASSO still fails to robustly perform correct identification when $n=150$. The poor performance of LASSO affects the bagging inclusion probability based LASSO estimator, which even underperforms STLS. For bagging inclusion probability based STLS, it has similar performance compared to STLS when the sample size is small, but performs much better with larger number of samples. When $n=150$, bagging inclusion probability based STLS performs perfect variable selection. 
    \item Model 3 generates a design matrix from the sparse physics model discovery setting. 
    We set $\gamma=50p\log(p)$ as a constant threshold for STLS, $\lambda=0.5$ for $\ell_1$ constraint in LASSO. For both bagging inclusion probability based LASSO and bagging inclusion probability based STLS, we subsample $50\%$ of data and set the inclusion probability threshold to be $0.8$. 
    In this setting, the terms in $X$ are dependent on the latent variables. This dependence structure violates many statistical conditions which are challenging for LASSO based estimators. 
    In Fig.~\ref{fig:fig:rel_freq} (i), (k), the LASSO based estimate of $\beta_{0,0}$ is consistently incorrect with $\ell_1$ constraints ranging from $[0.01, 100.0]$. For this setup, LASSO fails to identify any constant term from symbolic regression. Even if the bagging inclusion probability based method helps LASSO to improve the relative frequency for the other terms, but the identification of $\beta_{0,0}$ is still incorrect. 
    By contrast, STLS has good performances with increasing number of samples, and bagging inclusion probability based STLS significantly improves from STLS. Bagging inclusion probability based STLS behaves the best among these methods. 
\end{itemize}

\subsubsection{Robustness of hyper-parameters in varying sparse linear regression settings}

In practice, one may not have any prior knowledge about the sparsity level of a specific system.
We exemplify the idea of hyperparameter robustness in this experiment.  
For example in Model 1, we only know $p=30$, while $q$ can be an arbitrary value in the range $[0, 30]$. However, in non-asymptotic regimes, the LASSO type of variable selection methods are very sensitive to the choices of hyper-parameters. When $q$ is small (e.g. $q=1$), a large $\ell_1$ constraint is preferred; while, in contrast when $q\approx p$, a small $\ell_1$ constraint is preferred. It is important to understand if the hyperparameter setting can be generalized to practical  experiments. 

In the following, we perform the experiments similar to Model 1 with: (a) varying noise levels that $q=15, \sigma=[0.25, 0.5, 0.75, 1.0]$ and (b) varying number of active terms $\sigma=0.5, q=[1, 5, 10, 15]$. Different from many previous studies, we fix the hyperparameter setting, and understand how robust the hyperparameter is given the changes. We consistently fine tune threshold for all methods to obtain good empirical True discovery probability (ETDP), and compare their performances for empirical false discovery probability (EFDP). 

\begin{figure}[t]
    \centering
    \begin{overpic}[width=\textwidth]{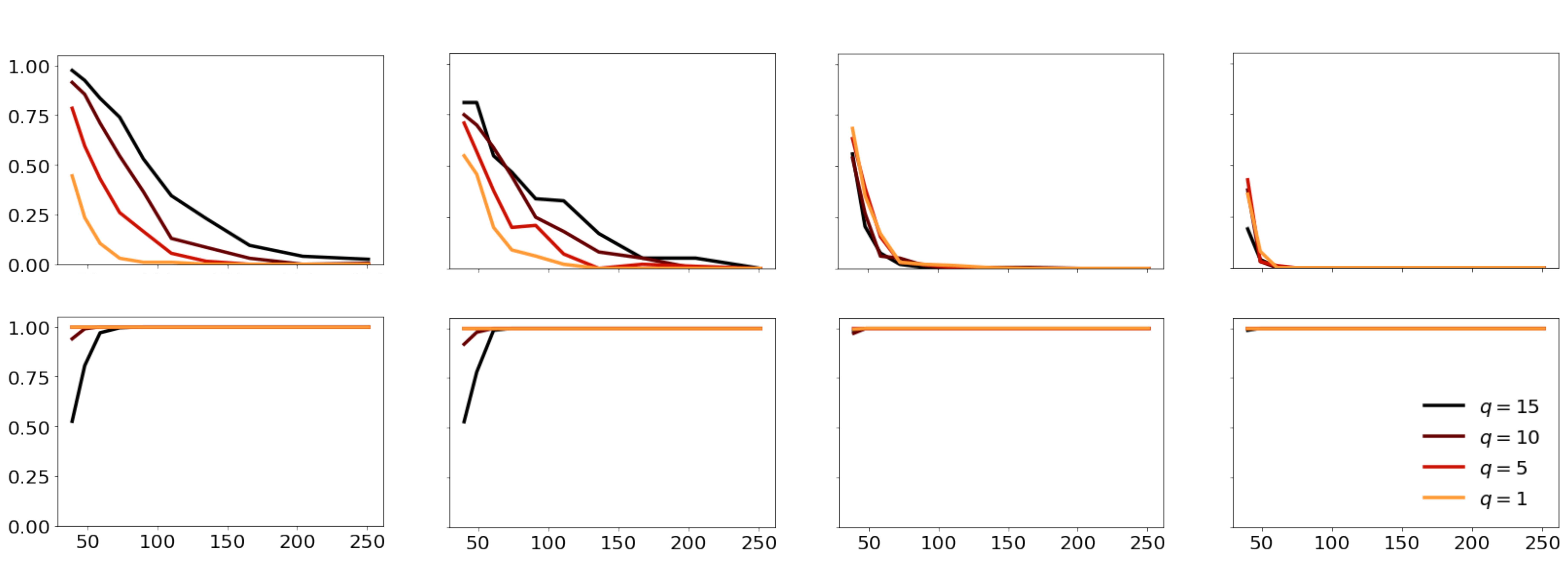}
    \put(10,33.0){\normalsize LASSO}
    \put(35,33.0){\normalsize BLASSO}
    \put(61,33.2){\normalsize STLS}
    \put(85.5,33.2){\normalsize BSTLS}
    \put(9,-2.0){\normalsize Sample size}
    \put(34,-2.0){\normalsize Sample size}
    \put(59,-2.0){\normalsize Sample size}
    \put(85,-2.0){\normalsize Sample size}
    \put(-2,22.0){\rotatebox{90}{\normalsize ETDP}}
    \put(-2,6.0){\rotatebox{90}{\normalsize EFDP}}
    \end{overpic}
    \vspace{0.1mm}
    \caption{Empirical false discovery probability and success true discovery probability with increasing number of samples under \textbf{different sparsity levels}. For better comparison, the hyperparameters of all methods are fine tuned for STDP that all non-zero coefficients are identified. We compare each method's performance via false discovery control. }
    \label{fig:q_various}
\end{figure}

\begin{figure}[t]
    \centering
    \begin{overpic}[width=\textwidth]{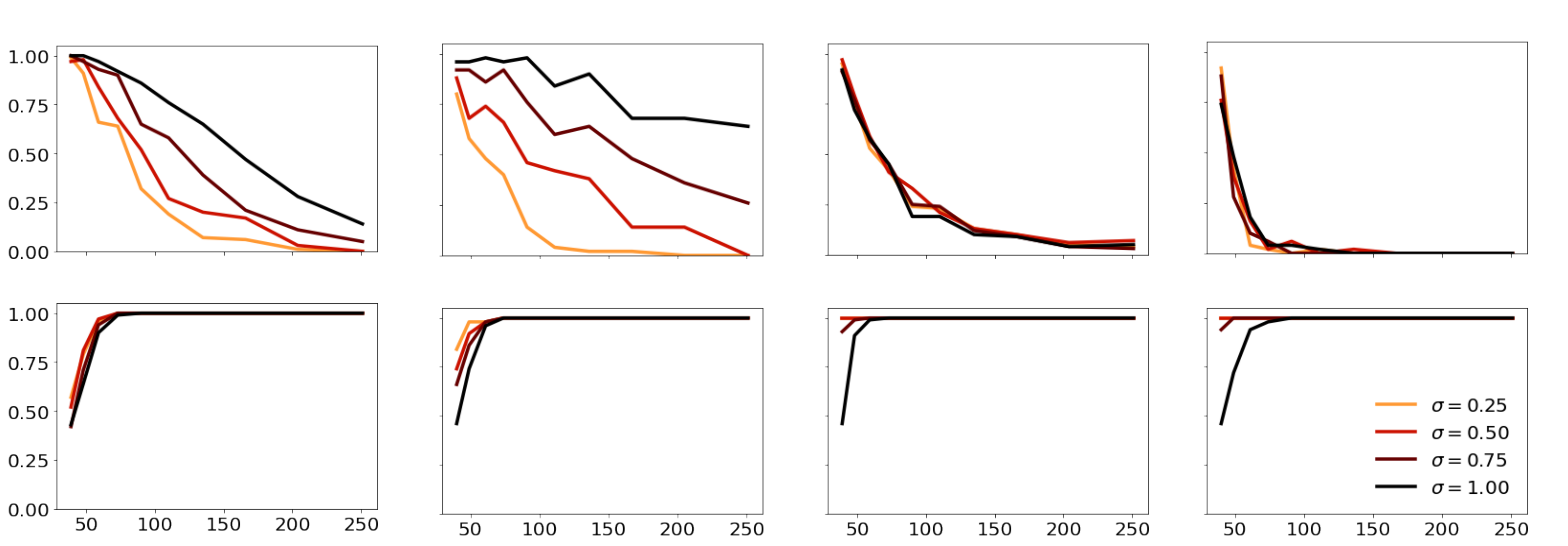}
    \put(10,33.0){\normalsize LASSO}
    \put(35,33.0){\normalsize BLASSO}
    \put(61.5,33.2){\normalsize STLS}
    \put(85.5,33.2){\normalsize BSTLS}
    \put(9,-2.0){\normalsize Sample size}
    \put(34,-2.0){\normalsize Sample size}
    \put(59,-2.0){\normalsize Sample size}
    \put(85,-2.0){\normalsize Sample size}
    \put(-2,22.0){\rotatebox{90}{\normalsize ETDP}}
    \put(-2,6.0){\rotatebox{90}{\normalsize EFDP}}
    \end{overpic}
    \vspace{0.1mm}
    \caption{Empirical false discovery probability and success true discovery probability with increasing number of samples under \textbf{various levels of noises}. For better comparison, the hyperparameters of all methods are tuned to control the worse case of ETDP is around $50\%$. We compare each method's performance via false discovery control.}
    \label{fig:noise_various}
\end{figure}
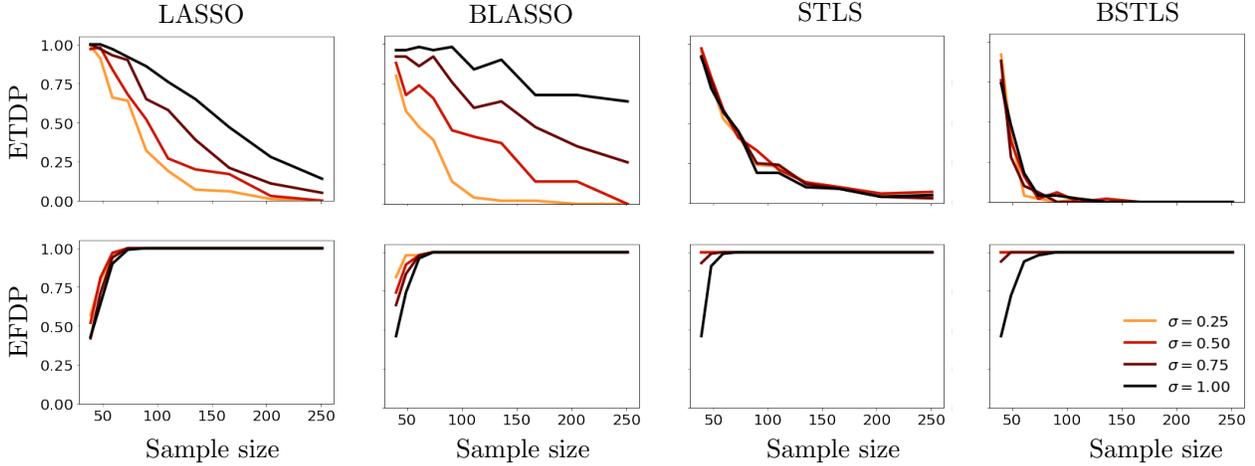

As shown in Fig.~\ref{fig:q_various} and Fig.~\ref{fig:noise_various}, we notice bagging inclusion probability based STLS is very robust to experimental settings. For both changing number of non-zero components and noise level, bagging inclusion probability based STLS performs similar in the control of false discovery probability. Even if bagging inclusion probability based STLS behaves slightly different in the empirical true discovery probability, the difference quickly vanishes after having relatively larger sample sizes (around $n=100$). STLS is also robust to changing experimental setting with relative worse performance comparing to bagging inclusion probability based methods. For example, in Fig.~\ref{fig:q_various} and Fig.~\ref{fig:noise_various}, the empirical FDP of bagging inclusion probability based methods perform similarly for both small noise (non-active terms) and large noise (non-active terms).

LASSO based methods all suffer from these changes that the false discovery controls perform very differently among these settings. Even if bagging inclusion probability based LASSO could help to control false discoveries comparing to LASSO, we could observe the false discovery curves behave very differently among these settings. For example, in Fig.~\ref{fig:q_various}, bagging inclusion probability based LASSO performs well when $q$ is large, but is relatively worse with higher sparsity rate. Similarly, in Fig.~\ref{fig:noise_various}, bagging inclusion probability based LASSO performs well  when $\sigma=0.25$, and behaves progressively worse with increasing noise levels. 
Furthermore, unlike STLS and bagging inclusion probability based STLS, the difference does not vanish even with relatively large sample sizes ($n=250$). 

\subsection{Lotka–Volterra model discovery via Ensemble SINDy}
\label{sec:physics_simulation}

The Lotka-Volterra model is frequently used to simulate the interaction between two competing groups $u$ and $v$ like predator-prey, chemical reactions, and economics. The model can be defined using the following set of differential equations: 
\begin{align}
    \dot{u} = \alpha u + \beta uv \\
    \dot{v} = \gamma v + \delta uv, 
\end{align}
where $\alpha=1.0, \beta=-0.1, \gamma=-1.5, \delta=0.075$. We set the initial conditions to $[u_0, v_0]=[10,5]$ with $\text{Lognormal}(0, 0.1)$ noise. 
The number of measurement data are uniformly sampled from $t\in[0, 24]$ with $n=[100, 200, 300, 400, 500]$. We normalize the measurement data via $u/\hat{\sigma}(u)$ and $v/\hat\sigma(v)$ where $\hat\sigma(\cdot)$ is the empirical standard deviation estimate. 
After normalization, the governing differential equation will have a different set of parameters  $\Tilde{\alpha}=1.0, \Tilde{\beta}=-0.68, \Tilde{\gamma}=-1.5, \Tilde{\delta}=0.82$. 
Compared to the synthetic simulation in~\ref{sec:simulation_model_setup}, the study in this subsection generates a time-series from a given initial condition $[u_0, v_0]$. This setting closely mimics the data collected by real-world sensors, and the temporal dependency structure presents a particularly challenging aspect.


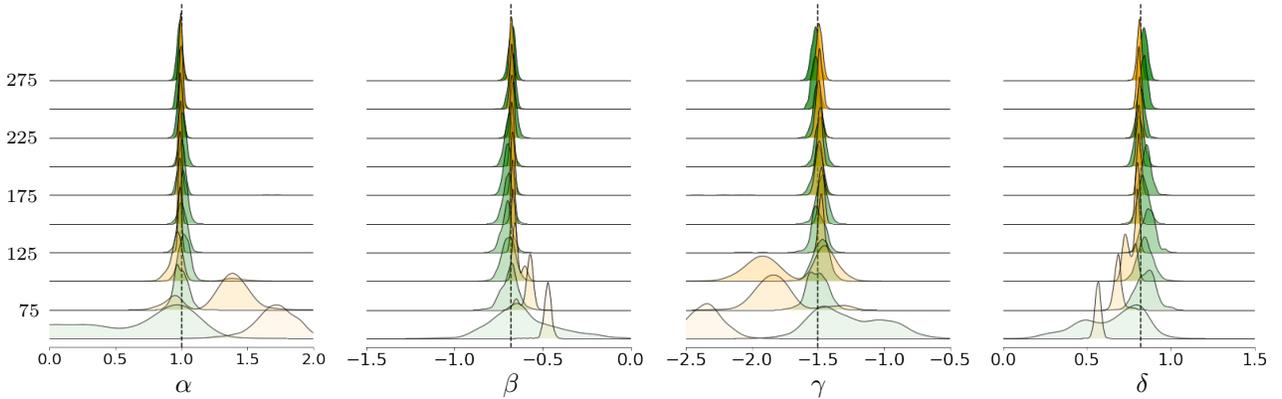
\begin{figure}[t]
    \centering
    \begin{overpic}[width=\textwidth]{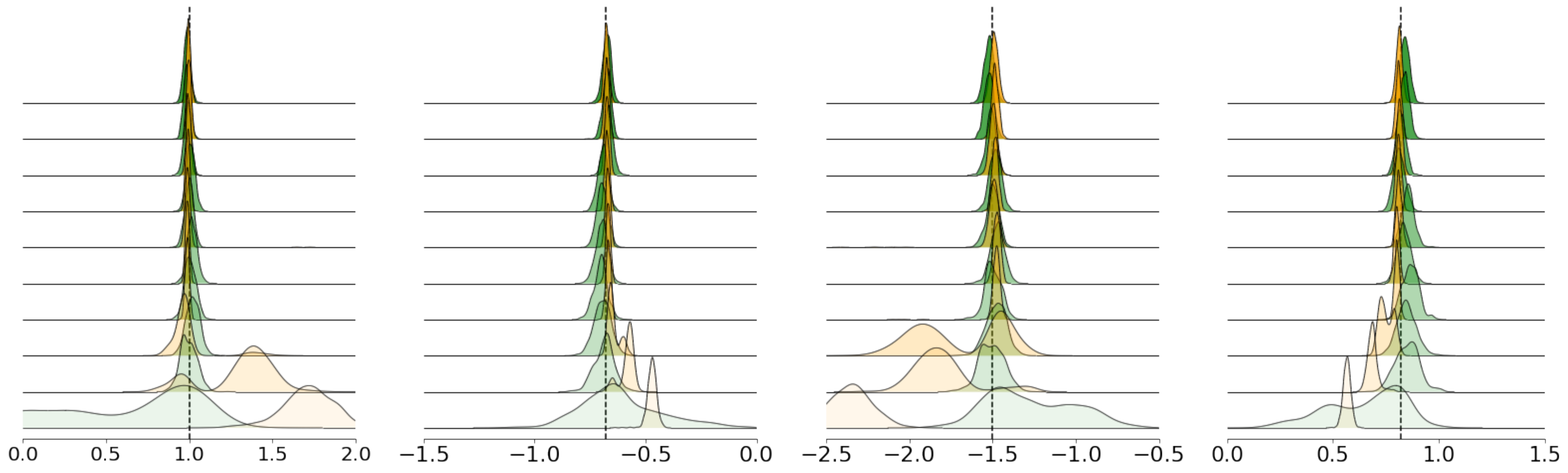}
    \put(11.6,-1.5){\normalsize $\alpha$}
    \put(38,-1.5){\normalsize $\beta$}
    \put(62.8,-1.5){\normalsize $\gamma$}
    \put(89,-1.5){\normalsize $\delta$}
    \put(-1,4.6){\scriptsize $75$}
    \put(-2,9.2){\scriptsize $125$}
    \put(-2,13.8){\scriptsize $175$}
    \put(-2,18.5){\scriptsize $225$}
    \put(-2,23.2){\scriptsize $275$}
    \end{overpic}
    \caption{Distributional approximation with uncertainty quantification of Bayesian SINDy (green-colored) and E-SINDy (orange-colored) to the ground-truth value (dotted line) for all active terms. }
    \label{fig:lotka}
\end{figure}

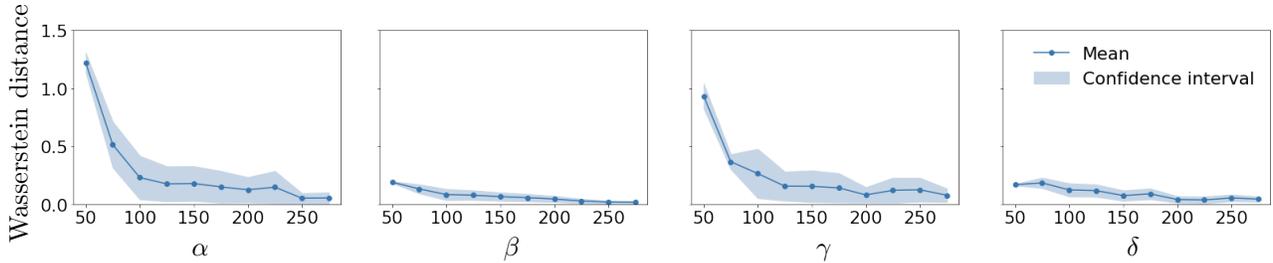
\begin{figure}[t]
    \centering
    \begin{overpic}[width=1.0\textwidth]{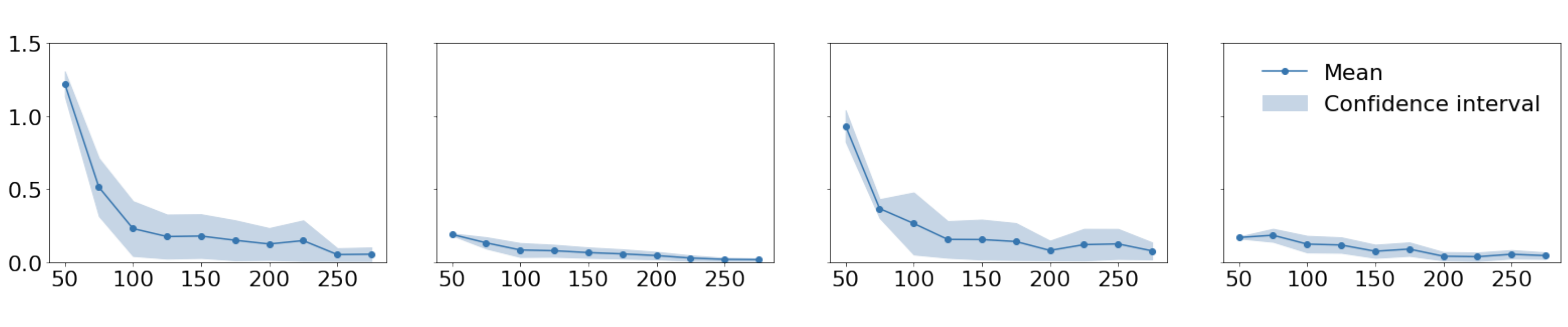}
    \put(-2,1.0){\rotatebox{90}{\normalsize Wasserstein distance}}
    \put(12.7,-0.0){\normalsize $\alpha$}
    \put(37.8,-0.0){\normalsize $\beta$}
    \put(63,-0.0){\normalsize $\gamma$}
    \put(88,-0.0){\normalsize $\delta$}
    \end{overpic}
    \caption{Convergence in Wasserstein-2 distance between E-SINDy and the Bayesian SINDy estimates for all four active indices. }
    \label{fig:w2_all}
\end{figure}

We visualize the distributional evolution for the four active coefficient estimates using E-SINDy and Bayesian SINDy~\citep{hirsh2022sparsifying} in Fig.~\ref{fig:lotka}. For all four indices ($\Tilde{\alpha} u, \Tilde\beta uv, \Tilde\gamma v, \Tilde\delta uv$), we observe the distributional approximation of E-SINDy would first transit from a wrong distribution (with biased mean, high variance) to a delta function concentrating around the true value. This observed behavior of E-SINDy makes it a suitable method for uncertainty quantification. While Bayesian SINDy can be computationally expensive, as shown in Fig.~\ref{fig:lotka}, it produces nearly unbiased distributional estimates even with very little data, which is remarkable for UQ purposes.
We further examine the convergence in Wasserstein between Bayesian SINDy and E-SINDy for all four active indices in Fig.~\ref{fig:w2_all}. For all indices, the Wasserstein distance converges quickly towards zero with larger number of samples.


\section{Conclusion and discussion}
\label{sec:discussion}

In this paper, we introduce the bagging inclusion probability-based sequentially thresholding least-squares estimation and establish the analysis of its statistical properties on variable selection as well as uncertainty quantification. 
On the one hand, for variable selection, bagging inclusion probability based STLS has the oracle property as stated in Thm.~\ref{thm:oracle_property}. Compared to STLS, bagging inclusion probability based STLS behaves better in theory, with the convergence of FDP and TDP being very fast with weaker assumptions. 
We study this difference through experiments in Sec.~\ref{sec:simulation}, which shows accelerated variable selection in various case studies. 
We note here that the error bound shown in this manuscript is not very tight. This is because we pay linearly with increasing model sizes $p$ in the empirical process with maximal of $n$ Binomial trails using union bounds. 
In practice, we observe that the errors are not growing linearly with $p$, but it seems to be nontrivial in theory. 
Besides, a non-asymptotic analysis of the mean squared error for bagging inclusion probability based STLS is not straightforward. Different inclusion probability threshold settings can lead to varying performances, and finding the optimal inclusion probability threshold can be challenging. The inclusion probability threshold may be able to control the false discovery rate through methods like Knockoffs~\citep{barber2015controlling,candes2018panning}. 
Therefore, we only provide qualitative results in this manuscript and leave more detailed quantitative descriptions for future works.

On the other hand, bagging inclusion probability based STLS also has a theoretical guarantee on variance estimation. As shown in Thm.~\ref{lem:uq_BIP}, we can use bootstrap replicates to approximate the variance of the estimator. 
In the experimental study with governing equation discovery (Sec.~\ref{sec:physics_simulation}), we visualize the distributional estimation in Fig.~\ref{fig:lotka} to understand the behavior of uncertainty quantification of the bagging inclusion probability based STLS estimator, which is important to understand for trustworthy implementation for prediction algorithms. 
An important aspect of our paper is to show the connection between Bayesian sparse inference via MCMC and bagging inclusion probability based STLS. 
Based on the rich history of understanding bootstrapping procedure as an approximate Bayesian inference~\citep{newton1994approximate,nie2022deep,rubin1981bayesian}, it is possible to understand bagging inclusion probability based STLS as an approximate Bayesian sparse inference. We show their equivalence by showing these two methods are asymptotically equivalent in Thm.~\ref{thm:w2_convergence}, yet Bayesian sparse inference via spike-and-slab prior has a very high computational requirement.

There are many interesting future extensions of our work. Given the computational efficiency and accuracy of ensemble based sparse regression methods, an avenue of ongoing research is to apply the technique to the broader class of sparse optimization problems like matrix completion and compressive sensing~\citep{zheng2018unified,candes2011robust,candes2006robust,candes2006stable}.
Second, it will be interesting to study the behavior of bootstrap sparse regression in high dimensional settings~\citep{karoui2016can,chernozhukov2017central}. 
Fasel et al.~\cite{fasel2022ensemble} empirically show that the E-SINDy algorithm with bagging inclusion probability is robust in high dimensional settings for PDE discovery.
This observation suggests the bootstrapping-based sparse regression method could mitigate some challenges reported in high dimensional settings, where it is generally difficult to have robust bootstrapping techniques~\citep{meinshausen2010stability,bach2008bolasso,chatterjee2011bootstrapping}. 
Finally, it will be exciting to extend the established theory to the different variants of ensembling techniques in E-SINDy~\citep{kaptanoglu2021pysindy}. This will be useful for practitioners to gain a better understanding of the different ensembling techniques in E-SINDy and their performance and applicability to different sparse model discovery settings. 

\section*{Acknowledgements}

The authors acknowledge support from the National Science Foundation AI Institute in Dynamic Systems (grant number 2112085).

\bibliographystyle{plain}
\bibliography{main.bbl}

\newpage
\appendix
\section{Notation}
\begin{tabular}{p{4.0cm}p{10.5cm}}
$\ve{X}$ & covariate input for linear model in $\real^{n\times p}$\\
$\ve{y}$ & label output for linear model in $\real^{n}$\\
$\beta$ & the unknown parameter of interest in $\real^p$\\
$\bar\beta$ & ordinary least-squares (OLS) estimate\\
$\epsilon$ & unobserved error follows distribution $P$\\
$I$ & All indices of unknown parameter where $I=\{1, ..., p\}$\\
$K_\beta$ & set of all non-active indices that $K_\beta = \{j\in I: \beta_j = 0\}$ \\
$J_\beta$ & set of all active indices that $J_\beta = \{j\in I: \beta_j \neq 0\}$ \\
$\hat{K}$ & thresholding estimator of the index set of zero components $K_\beta$ \\
$\Theta_1$ & set of all sparse solutions $\Theta_1 = \{\beta\in\real^p:K_\beta\neq \varnothing\}$\\
$\Theta_2$ & set of all non-sparse solutions $\Theta_2 = \real^p \backslash \Theta_1$ \\
$n$ & number of sample data\\
$p$ & size of linear model\\
$q$ & number of sparse variables that $q=\text{card}(J)$\\

\end{tabular}

\section{Proof of Lemma~\ref{lem:ols_tail}}
Before showing the Lemma, we first introduce the following regularity conditions. 

\vspace{3mm}
D.1 $\E{\epsilon}=0$, $Var(\epsilon)=\sigma^2<\infty$, and $\E{\epsilon^4}<\infty$ where $\epsilon\sim P$.  \label{lem:assumption}

D.2 $\rho_1 > 0, \max_i (X_i^TX_i)=O(p)$, and $p/(n\rho_1)=o(1)$. 
\vspace{3mm}

\begin{proof}
We first show Eqn.~\eqref{lemeq:normality}. We could see that $\bar\beta$ is an unbiased estimate that 
\begin{align}
    \label{pf:unbias_normality}
    \Ep{\epsilon}{a^T(\bar\beta-\beta)} = \Ep{\epsilon}{a^T(X^TX)^{-1}X^T\epsilon} = 0.
\end{align}

We write
\begin{align}
    a^T(\bar\beta-\beta)=a^T(\ve{X}^T\ve{X})^{-1}\ve{X}^T\ve{\epsilon} = \sum_{i=1}^n \alpha_i\epsilon_i,
\end{align}
where $\alpha_i = a^T(\ve{X}^T\ve{X})^{-1}X_i$. 

Showing~\eqref{lemeq:normality} is equivalent to show $\sum_{i=1}^n\alpha_i\epsilon_i \to \mathcal{N}\lrp{0, \sigma^{2}a^T(X^TX)^{-1}a}$. 

Due to~\eqref{pf:unbias_normality}, we see $\E{\sum_{i=1}^n \alpha_i\epsilon_i}=0$. 

Consider the variance of $\alpha_i\epsilon_i$, 
\begin{align}
    var(\alpha_i\epsilon_i) = \sigma^2a^T(X^TX)^{-1}X_iX_i^T(X^TX)^{-1}a
\end{align}

We have
\begin{align}
    \sum_{i=1}^n var(\alpha_i\epsilon_i) = \sigma^2a^T(X^TX)^{-1}a
\end{align}

By Lindeberg-Feller Central Limit Theorem, we can show the weak convergence to normal distribution. We apply Lindeberg-Feller CLT, we need to verify the Lindeberg condition that 
\begin{align}
    \sum_{i=1}^n \E{|\alpha_i\epsilon_i|^2\ind{|\alpha_i\epsilon_i|\geq\delta}}\to 0, \;\;\;\;\forall\delta>0.
\end{align}

We know
\begin{align}
    \sum_{i=1}^n \E{|\alpha_i\epsilon_i|^2\ind{|\alpha_i\epsilon_i|\geq\delta}} &\leq \sum_{i=1}^n \E{\alpha_i^2\epsilon_i^2} \\
    &= \sigma^2a^T(X^TX)^{-1}a\\
    &\leq \frac{\sigma^2}{n\rho_1} \\
    &= o(1).
\end{align}

Since Lindeberg condition is satisfied, using Lindeberg-Feller CLT and Slutsky's theorem, we can conclude~\eqref{lemeq:normality}. 

Using sub-Gaussian variable result, by setting the right hand side as $\delta$, we could exactly see the high-probability tail form as~\eqref{lemeq:tail_bound}. 
\end{proof}
\section{Proof of Lemma~\ref{lem:tlse_oracle}}
Say $q_{K_\beta}$ is the number of active indices in $\beta$, $n$ is the number of sample, and $p$ is the model size. We require the following regularity conditions to show this Lemma. 

\vspace{3mm}
(a) $\frac{q_{K_\beta}}{n}=o(1)$. 

(b) $\lrp{\frac{n^{1/2}\rho_1^{1/2}|\beta_{(1)}|}{\sigma}-\sqrt{2\log(n)}-\sqrt{2\log(p-q_{K_\beta})}}^2\to\infty$. 
\vspace{3mm}

Condition (a) requires the sample size goes to infinity. 
Condition (b) is mild since as $n\to\infty$, $n^{1/2}-\sqrt{\log(n)}\to\infty$. 
Practically, as long as $|\beta_{(1)}|$ is a constant which does not scale with $\frac{1}{n}$, condition (b) should hold in general. 

\begin{proof}



To show $Pr(\hat{K}=K_\beta)=1$, we first consider $Pr(K_\beta\subseteq\hat{K})$. 
We can bound the chance of false discovery that and by setting $\delta=\frac{1}{n}$, 
\begin{align}
    Pr\lrp{\max_{j\in K_\beta} |\bar\beta_j| \geq\sqrt{2\sigma^2\log(1/\delta)(X_j^TX_j)^{-1}}} &\leq q_{K_\beta} Pr\lrp{|\bar\beta_1| \geq \sqrt{2\sigma^2\log(1/\delta)(X_j^TX_j)^{-1}}} \\
    &\leq \frac{q_{K_\beta}}{n}
\end{align}

From (a) we know $\frac{q_{K_\beta}}{n}=o(1)$. Therefore, 
\begin{align}
    Pr\lrp{|\bar\beta_j| \leq \sqrt{2\sigma^2\log(1/\delta)(X_j^TX_j)^{-1}}\text{ for all }j\in K_\beta} \to 1.
\end{align}

\begin{align}
    Pr\lrp{K_\beta \subseteq \hat{K}} \to 1.
\end{align}

From the other side, set $\beta_{(1)}$ to be the active coefficient with minimal magnitude. We could upper bound the probability that the minimum least-squares estimate of an active index will below the threshold that
\begin{align}
    Pr\lrp{\min_{j\in J_\beta}|\bar\beta_j| \leq \sqrt{2\sigma^2\log(1/\delta)(X_j^TX_j)^{-1}}} &\leq Pr\lrp{\min_{j\in J_\beta}|\beta_j|-|\bar\beta_j-\beta_j| \leq \sqrt{2\sigma^2\log(1/\delta)(X_j^TX_j)^{-1}}} \\
    &\leq Pr\lrp{\min_{j\in J_\beta}|\beta_j|-|\bar\beta_j-\beta_j| \leq \sqrt{2\sigma^2\log(1/\delta)(X_j^TX_j)^{-1}}} \\
    &\leq Pr\lrp{\max_{j\in J_\beta}|\bar\beta_j-\beta_j| \geq |\beta_{(1)}|- \sqrt{2\sigma^2\log(1/\delta)(X_j^TX_j)^{-1}}} \\
    &= 2Pr\lrp{\max_{j\in J_\beta}\lrp{\bar\beta_j-\beta_j} \geq |\beta_{(1)}|- \sqrt{2\sigma^2\log(1/\delta)(X_j^TX_j)^{-1}}}
\end{align}

Say $G_j = \frac{\bar\beta_j-\beta_j}{\sigma\sqrt{X_j^TX_j}^{-1}}$, we know $G_j\sim\mathcal{N}(0, 1)$. Therefore, 
\begin{align}
    Pr\lrp{\max_{j\in J_\beta}\lrp{\bar\beta_j-\beta_j} \geq |\beta_{(1)}|- \sqrt{2\sigma^2\log(1/\delta)(X_j^TX_j)^{-1}}} &= Pr\lrp{\max_{j\in J_\beta}G_i \geq \frac{|\beta_{(1)}|}{\sigma\sqrt{(X^T_jX_j)^{-1}}}- \sqrt{2\log(1/\delta)}} \\
    &\leq Pr\lrp{\max_{j\in J_\beta}G_i \geq \frac{n^{1/2}\rho_1^{1/2}|\beta_{(1)}|}{\sigma}- \sqrt{2\log(1/\delta)}}
\end{align}

Using the maximal result of sub-Gaussian random variables, we see
\begin{align}
    &Pr\lrp{\max_{j\in J_\beta}G_i -\sqrt{2\log(p-q_{K_\beta})}\geq \frac{n^{1/2}\rho_1^{1/2}|\beta_{(1)}|}{\sigma}- \sqrt{2\log(1/\delta)}-\sqrt{2\log(p-q_{K_\beta})}} \\
    &\qquad\qquad\qquad\qquad\qquad\qquad\qquad\leq\exp\lrp{-\frac{1}{2}\lrp{\frac{n^{1/2}\rho_1^{1/2}|\beta_{(1)}|}{\sigma}- \sqrt{2\log(1/\delta)}-\sqrt{2\log(p-q_{K_\beta})}}^2} 
\end{align}

Summarizing all the terms and let $\delta=\frac{1}{n}$, we know 
\begin{align}
    Pr\lrp{\min_{j\in J_\beta}|\bar\beta_j| \leq \sqrt{2\sigma^2\log(1/\delta)(X_j^TX_j)^{-1}}} &\leq 2
    \exp\lrp{-\frac{1}{2}\lrp{\frac{n^{1/2}\rho_1^{1/2}|\beta_{(1)}|}{\sigma}- \sqrt{2\log(n)}-\sqrt{2\log(p-q_{K_\beta})}}^2} 
\end{align}

From (b) we know $\lrp{\frac{n^{1/2}\rho_1^{1/2}|\beta_{(1)}|}{\sigma}-\sqrt{2\log(n)}-\sqrt{2\log(p-q_{K_\beta})}}\to\infty$, therefore

\begin{align}
    Pr\lrp{\min_{j\in J_\beta}|\bar\beta_j|\leq \sqrt{2\sigma^2\log(1/\delta)(X_j^TX_j)^{-1}}} \to 0.
\end{align}

\end{proof}
\section{Proof Theorem \ref{thm:oracle_property}}
\label{app:proof_theorem}
\begin{proof}
\textbf{Part (a) }

First, we know $\bar\beta_j^b$ is just the Bootstrap replicate of OLS estimator. From the assumption of the tail behavior, we see that $\bar\beta$ follows
\begin{align}
    \bar\beta &= \lrp{X^TX}^{-1}X^Ty \\
    &= \lrp{X^TX}^{-1}X^T\lrp{\beta X+\epsilon} \\
    &= \beta + \lrp{X^TX}^{-1}X^T \eta \\
    &\sim\mathcal{N}\lrp{\beta, \sigma^2\lrp{X^TX}^{-1}}
\end{align}

Therefore, from the sub-Gaussian tail behavior, by taking $a$ as the standard basis with index $j$, there exists some $\delta>0$ that
\begin{align}
    Pr\lrp{|\bar\beta_j-\beta|>\sigma\sqrt{2\log(1/\delta)\lrp{X^TX}^{-1}}} \leq 2\delta.
\end{align}




Similarly from the sub-Gaussian tail, we could see that there exists some other $\delta>0$ that
\begin{align}
    Pr\lrp{|\bar\beta_j^b-\beta|>\sigma\sqrt{2\log(1/\delta)\lrp{X^{b^T}X^b}^{-1}}} \leq 2\delta.
\end{align}




The quantity $\ind{|\bar\beta^b_j| > \sigma\sqrt{2\log(1/\delta)(X_j^{bT}X_j^b)^{-1}}}$ only takes $\{0,1\}$ which is a Bernoulli random variable. Therefore, we can consider a Bernoulli process $B_j^b\sim Bernoulli(2\delta)$. We know that $\mathbf{B}_j = \sum_{b=1}^n B^b_j \sim Binomial(n, 2\delta)$. We can compute
\begin{align}
    \label{eqn:tpp_binomial}
    Pr\lrp{\max_{j\in K_\beta} \frac{1}{n}\sum_{b=1}^n \ind{|\bar\beta^b_j| > 2\sigma\sqrt{2\log(1/\delta)(X_j^{bT}X_j^b)^{-1}}} > p_c} &\leq Pr\lrp{\max_{j\in K_\beta}\sum_{b=1}^n B_j^b > np_c} \\
    &= Pr\lrp{\max_{j\in K_\beta} \mathbf{B}_j > np_c} \\
    &= Pr\lrp{\bigcup_{j\in K_\beta} (\mathbf{B}_j > np_c)} \\
    &\leq (p-q)Pr\lrp{\mathbf{B}_1 > np_c}. 
\end{align}

Here we use $\mathbf{B}_1$ because any $j\in K_\beta$ is the same after applying the Union bound. 

From Eqn. \eqref{eqn:tpp_binomial}, we cast the original problem into an empirical process with Binomial trails. Therefore, we apply the following concentration inequalities to bound this probability.  


\textit{(a.1) Chernoff bound}

Since $B_j^b\in [0, 1]$, from Chernoff Upper Tail, and suppose $p_c \geq 2\delta$, we know
\begin{align}
    Pr\lrp{\mathbf{B}_j \geq np_c} \leq \exp\lrp{-\frac{2n\delta(p_c/2\delta-1)}{3}}
\end{align}

By setting $\delta=\frac{1}{n}$, 
\begin{align}
    Pr\lrp{\mathbf{B}_j \geq np_c} \leq \exp\lrp{\frac{2}{3}-\frac{np_c}{3}}. 
\end{align}


Therefore, with condition (C.1)
\begin{align}
    \label{eqn:tpp_convergence}
    Pr\lrp{\max_{j\in K_\beta} \frac{1}{n}\sum_{b=1}^n \ind{|\bar\beta^b_j| > 2\sigma\sqrt{2\log(1/\delta)(X_j^{bT}X_j^b)^{-1}}} > p_c} &\leq (p-q)\exp\lrp{\frac{1}{3}-\frac{np_c}{3}}\to0.
\end{align}


This means that 
\begin{align}
    Pr\lrp{\frac{1}{n}\sum_{b=1}^n \ind{|\bar\beta^b_j| > 2\sigma\sqrt{2\log(1/\delta)(X_j^{bT}X_j^b)^{-1}}} < p_c\;for\;all\;j\in K_\beta} \to 1.
\end{align}

Eqn. \eqref{eqn:tpp_convergence} shows both (i) one side of (iv) in the Theorem. 

To see the bound is tight, we apply another bound which obtains a similar rate of convergence. 

\textit{(a.2) Hoeffding bound}

Since $B_j^b\in [0, 1]$, from Hoeffding's bound, we know
\begin{align}
    Pr\lrp{\mathbf{B}_j -2n\delta \geq np_c-2n\delta} &\leq \exp\lrp{-\frac{2(np_c-2n\delta)^2}{n}} \nonumber\\
    &=\exp\lrp{-\frac{2(np_c-2)^2}{n}}
\end{align}

Therefore, with condition (C.1), we have
\begin{align}
    Pr\lrp{\max_{j\in K_\beta} \frac{1}{n}\sum_{b=1}^n \ind{|\bar\beta^b_j| > 2\sigma\sqrt{2\log(1/\delta)(X_j^{bT}X_j^b)^{-1}}} > p_c} &\leq q\exp\lrp{-\frac{2(np_c-2)^2}{n}}\to0.
\end{align}

This means that 
\begin{align}
    Pr\lrp{\frac{1}{n}\sum_{b=1}^n \ind{|\bar\beta^b_j| > \sigma\sqrt{2\log(1/\delta)(X_j^{bT}X_j^b)^{-1}}} < p_c\;for\;all\;j\in K_\beta} \to 1.
\end{align}

\textbf{Part (b) } 

Without loss of generality, we set $|\beta_1|=\min_{j\in J_\beta} |\beta_j|$ is the smallest. 


\begin{align}
    &Pr\lrp{\min_{j\in J_\beta} \frac{1}{n}\sum_{b=1}^n \ind{|\bar\beta^b_j| > 2\sigma\sqrt{2\log(1/\delta)(X_j^{bT}X_j^b)^{-1}}} < p_c} \\
    &\leq Pr\lrp{\bigcup_{j\in J_\beta} \lrp{\frac{1}{n}\sum_{b=1}^n \ind{|\bar\beta^b_j| > 2\sigma\sqrt{2\log(1/\delta)(X_j^{bT}X_j^b)^{-1}}} < p_c}} \\
    &\leq q Pr\lrp{\frac{1}{n}\sum_{b=1}^n \ind{|\bar\beta^b_j| > 2\sigma\sqrt{2\log(1/\delta)(X^{bT}_jX^b_j)^{-1}}} < p_c} \\
    &\leq q Pr\lrp{\frac{1}{n}\sum_{b=1}^n \ind{\frac{|\bar\beta^b_j|}{\sigma\sqrt{(X^{bT}_jX^b_j)^{-1}}} > 2\sqrt{2\log(1/\delta)}} < p_c} 
\end{align}

Consider the quantity on the left hand side $\frac{|\bar\beta^b_j|}{\sigma\sqrt{(X_j^{bT}X_j^b)^{-1}}}$. We know that the bootstrap replicate $\bar\beta^b_j\sim\mathcal{N}\lrp{\bar\beta_j, \sigma\sqrt{(X_j^{bT}X_j^b)^{-1}}}$. 
Therefore, $\frac{|\bar\beta^b_j|}{\sigma\sqrt{(X_j^{bT}X_j^b)^{-1}}}\sim\mathcal{N}\lrp{\frac{|\bar\beta_j|}{\sigma\sqrt{X_j^{bT}X_j^b)^{-1}}}, 1}$. 

The mean of the above quantity is lower bounded by
\begin{align}
    \E{\frac{|\bar\beta^b_j|}{\sigma\sqrt{(X_j^{bT}X_j^b)^{-1}}}} &= \frac{|\bar\beta_j|}{\sigma\sqrt{(X_j^{bT}X_j^b)^{-1}}} \\
    &\geq \frac{\sqrt{n}|\bar\beta_j|}{\sigma\sqrt{(r_0\rho_1)^{-1}}} \\
    &\geq \frac{\sqrt{n}\lrp{|\beta_{1}|-\max_{j\in J_\beta}|\bar\beta_j-\beta_j|}}{\sigma\sqrt{(r_0\rho_1)^{-1}}} 
\end{align}



\textbf{Scenario 1: when we have large sample size. }

Suppose the sample size is relatively large enough, resulting $\frac{\sqrt{n}\lrp{|\beta_{1}|-\max_{j\in J_\beta}|\bar\beta_j-\beta_j|}}{\sigma\sqrt{(r_0\rho_1)^{-1}}} \geq 2\sqrt{2\log(n)}$.
This condition is mild since the $\sqrt{n}$ rate on the left hand side dominates the $\sqrt{\log{n}}$ rate on the right hand side. It is also reasonable to assume that $|\beta_1|, |\bar\beta_j-\beta_j|, \sigma, \sqrt{(r_0\rho_1)^{-1}}$ are constants. 

From this condition, we know that 
\begin{align}
    Pr\lrp{\ind{|\bar\beta^b_j| > 2\sigma\sqrt{2\log(1/\delta)(X_j^{bT}X_j^b)^{-1}}}} \geq 0.5.
\end{align}


Therefore, consider a Bernoulli process $B_j^b\sim Bernoulli(0.5)$. We know that $\mathbf{B}_j = \sum_{b=1}^n B^b_j \sim Binomial(n, 0.5)$. We can compute using Chernoff Lower Tail 
\begin{align}
    Pr\lrp{\mathbf{B}_j \leq np_c} \leq \exp\lrp{-\frac{n(1-2p_c)^2}{6}}
\end{align}

Therefore, with condition (A.3)
\begin{align}
    Pr\lrp{\max_{j\in J_\beta} \frac{1}{n}\sum_{b=1}^n \ind{|\bar\beta^b_j| > 2\sigma\sqrt{2\log(1/\delta)(X_j^{bT}X_j^b)^{-1}}} < p_c} &\leq p\exp\lrp{-\frac{n(1-2p_c)^2}{6}}\to0.
\end{align}

\textbf{Scenario 2: before we have that large amount of samples. }

It is clear that scenario 1 holds with large sample sizes. However, what happens if we do not have that large amount of sample? For example, the data may have low signal-to-noise ratio with very limited sample size; or $\beta_1$ is considerably small compared to other $\beta_j$ terms ($j\in J_\beta$), but the effect still cannot be neglected.

The following analysis will consider this situation under these concerns. We achieve the following analysis by putting a stronger condition on the bounds of Gaussian tails for all indices $j\in J_\beta$.

From (A.5) and Theorem 3 in \citep{zhang2020non}, we see there exist constants $c,c',C>0$ that
\begin{align}
    \label{eqn:lower_bd_pc}
    Pr\lrp{\bar\beta^b_j - \beta_j \geq 2\sigma\sqrt{2\log(1/\delta)(X_j^{bT}X_j^b)^{-1}}-\beta_j} &\geq c\exp\lrp{-C\frac{\lrp{2\sigma\sqrt{2\log(1/\delta)(X_j^{bT}X_j)^{-1}}-\beta_j}^2}{4\sigma^2\log(1/\delta)(X_j^TX_j)^{-1}}} \\
    \boxed{Let\;s=2\sqrt{2\log(1/\delta)(X_j^{bT}X_j^b)^{-1}}}\;\;\;\;\;\;\;\;\;\;\;\;\;\;\;&= c\exp\lrp{-C\frac{\lrp{s-\beta_j}^2}{2s^2}} \\
    \boxed{From\;\beta_j\in(0,s)}\;\;\;\;\;\;\;\;\;\;\;\;\;\;\;&\geq c\exp\lrp{-C}.
\end{align}

Therefore, we can always choose a constant threshold $p_c < \min_{j\in J_\beta} c\exp\lrp{\frac{-C\lrp{2\sigma\sqrt{2\log(1/\delta)(X_j^TX_j)^{-1}}-\beta_j}^2}{4\sigma^2\log(1/\delta)(X_j^TX_j)^{-1}}}$. Notice here $p_c$ will not scale to $0$ due to the lower bound in Eqn. \eqref{eqn:lower_bd_pc}. 

Then, we could set $p_{j_{min}}=\min_{j\in J_\beta} c\exp\lrp{\frac{-C\lrp{2\sigma\sqrt{2\log(1/\delta)(X_j^TX_j)^{-1}}-\beta_j}^2}{4\sigma^2\log(1/\delta)(X_j^TX_j)^{-1}}}$, and treat $B^b_j\sim Bernoulli(p_{j_{min}})$. We know that $\mathbf{B}_j = \sum_{b=1}^n B^b_j \sim Binomial(n, p_{j_{min}})$. We can compute using Chernoff Lower Tail 
\begin{align}
    Pr\lrp{\mathbf{B}_j \leq np_c} &\leq \exp\lrp{-\frac{np_{j_{min}}(1-p_c/p_{j_{min}})^2}{3}} \\
    &= \exp\lrp{-\frac{np_c^2}{3p_{j_{min}}}+\frac{2np_c}{3}-\frac{np_{j_{min}}}{3}}\\
    &= \exp\lrp{\lrp{-\frac{p_c^2}{3p_{j_{min}}} + \frac{2p_c}{3} - \frac{np_{j_{min}}}{3}}}
\end{align}

Even if the rate of convergence is much slower comparing to the large sample condition, it is still exponential with respect to sample size $n$. 
We see if $p_c\ll p_{j_{min}}$, we can bound as $Pr\lrp{\mathbf{B}_j \leq np_c} \leq \exp\lrp{-\frac{np_{j_{min}}}{3}}$. However, the condition $p_c\ll p_{j_{min}}$ is typically equivalent to the large sample condition (as in Condition 1). 

\end{proof}
\section{Proof of Lemma 4.2}
\begin{proof}
Consider this definition of $\Delta$
\begin{align}
    \Delta &= \underbrace{\min_{j\in J_\beta} \frac{1}{n}\sum_{b=1}^n \ind{|\bar\beta^b_j| > \sigma\sqrt{2\log(1/\delta)(X_j^{bT}X_j^b)^{-1}}}}_{\mathcal{I}_1} - \underbrace{\max_{k\in K_\beta} \frac{1}{n}\sum_{b=1}^n \ind{|\bar\beta^b_k| > \sigma\sqrt{2\log(1/\delta)(X_k^{bT}X_k^b)^{-1}}}}_{\mathcal{I}_2}
\end{align}

Given the assumptions, conditions (C.2) and proof in Thm.~\ref{thm:oracle_property}, c, we know the random variables in $\mathcal{I}_1$ and $\mathcal{I}_2$ can be lower/upper bounded by $Binomial(n, 0.5)$ and $Binomial(n, 1/n)$ respectively. 

Therefore, in $\mathcal{I}_1$, using Chernoff bound, we see with probability at least $1-qe^{-n\lrp{\frac{1}{4}-\frac{\epsilon}{2}-\frac{1}{2n}}^2/2}$, 
\begin{align}
    \mathcal{I}_1 \geq 0.25+\frac{\epsilon}{2}+\frac{1}{2n}.
\end{align}

Similarly, in $\mathcal{I}_2$, using Chernoff bound, we see with probability at least $1-(p-q)e^{-n\lrp{\frac{1}{4}-\frac{\epsilon}{2}-\frac{1}{2n}}^2/2}$, 
\begin{align}
    \mathcal{I}_2 \leq 0.25-\frac{\epsilon}{2}+\frac{1}{2n}.
\end{align}

Consider the joint event with an union bound, we know that $\Delta \geq \epsilon$ holds with probability at least 
\begin{align}
\lrp{1-e^{-n\lrp{\frac{1}{4}-\frac{\epsilon}{2}-\frac{1}{2n}}^2/2}}^q\lrp{1-e^{-n\lrp{\frac{1}{4}-\frac{\epsilon}{2}-\frac{1}{2n}}^2/2}}^{p-q} &=\lrp{1-e^{-n\lrp{\frac{1}{4}-\frac{\epsilon}{2}-\frac{1}{2n}}^2/2}}^p \\
&\geq 1-pe^{-n\lrp{\frac{1}{4}-\frac{\epsilon}{2}-\frac{1}{2n}}^2/2}.
\end{align}
\end{proof}
\section{Proof Theorem \ref{lem:uq_BIP}}
\label{app:proof_uq_BIP}
In the asymptotic settings, we do not require further distributional assumptions (as shown in C.1). However, since we do not have non-asymptotic Gaussian tail behaviors, the threshold for subset estimation is alternatively determined via the standard deviation estimate with a factor $\gamma$ as in~\citep{giurcanu2016thresholding}. In specific, \begin{align}
    \hat K_{BIP} = \left\{j\in I: \frac{1}{n}\sum_{b=1}^n \ind{|\bar\beta^b_j| > \gamma\bar{\sigma}^b_{jj}} \leq p_c\right\}. 
\end{align}

We further require the following conditions: 

(c) The min eigenvalue $\rho_1$ and max eigenvalue $\rho_2$ satisfies $\frac{\rho_1\rho_2^{-1}\gamma^2-q_{K_\beta}}{q^{1/2}_{K_\beta}} \to \infty$. 

(d) $\frac{\rho_1\rho_2^{-1}\lrp{n^{1/2}\sigma^{-1}\rho_1^{1/2}\min_{j\in J_\beta} |\beta_j|-\gamma}^2-q}{q^{1/2}} \to \infty$

\begin{proof}







First, by the construction of $\hat\beta^*$, it performs least-square estimation based on the bagging inclusion probability. The Bootstrap replicates are generated from resampling of the original data, which can be traversed. Therefore, when $m\to\infty$, the conditional estimation of $\hat{K}_{BIP}^\infty\mid Y$ will be a fixed set of indices. 

Then, we show that $\hat\sigma_m^{^*2}\xrightarrow[]{P}\sigma^2$. Denote the event $\mathcal{E}$ which includes all cases when $\{\hat{K}_{BIP}^\infty=\hat{K}=K_\beta\mid Y\}$. This event describes when both BIP and TLS are performing a correct subset estimation. We have
\begin{align}
    \lim_{n,m\to\infty}\left\{P\lrp{|\hat\sigma_m^{^*2}-\sigma^2|>\epsilon}\mid Y\right\} &\leq \lim_{n,m\to\infty}\left\{P\lrp{|\hat\sigma_m^{^*2}-\sigma^2|>\epsilon\mid\mathcal{E}}P(\mathcal{E}) + P\lrp{|\hat\sigma_m^{^*2}-\sigma^2|>\epsilon\mid\mathcal{E}^c}P(\mathcal{E}^c)\right\} \\
     &= 0+o_P(1) = o_P(1).
\end{align}

We see the probability $P\lrp{|\hat\sigma_m^{^*2}-\sigma^2|>\epsilon\mid\mathcal{E}}$ converges to zero by Thm.~2.2 (b)~\citep{freedman1981bootstrapping}. For the other term, notice that $P(\mathcal{E})$ in the asymptotic regime ($n\to\infty$), with all the regularity conditions, we have $P(\hat{K}\neq K_\beta)= o_P(1)$ and $P(\hat{K}_{BIP}= K_\beta)\leq o_P(1)$. By using the union bound, we see $P(\mathcal{E}^c)= o_P(1)$, which gives us $\hat\sigma_m^{^*2}=\sigma^2+o_P(1)$. In general, most of the convergence in probability results in the Theorem can be shown in this way due to $P(\mathcal{E}^c)= o_P(1)$. We constraint first within the event $\mathcal{E}$ in the following analysis. 

Conditioning on the event $\mathcal{E}$, we have $\hat{K}_{BIP}^\infty=\hat{K}=K_\beta$. By the Slutsky's theorem, it suffices to show that the law of $\hat{s}^{-1/2}a^T(\hat\beta^*-\hat\beta\mid Y)$ is consistent. 

\begin{align}
    \hat{s}^{-1/2}a^T(\hat\beta^*-\hat\beta) &= \hat{s}^{-1/2}a_{\hat J}^T((X_{\hat J}(m)^TX_{\hat J}(m))^{-1}X_{\hat J}^T(m)(X_{\hat J}(m)\hat\beta+\hat\epsilon^*(m))-\hat\beta) \\
    &= \hat{s}^{-1/2}a_{\hat J}^T((X_{\hat J}(m)^TX_{\hat J}(m))^{-1}X_{\hat J}(m)^T\hat\epsilon^*(m) \\
    &= \sum_{i=1}^m \hat\alpha_i\hat\epsilon^*_i,
\end{align}
where $\hat\alpha_i=\hat{s}^{-1/2}a_{\hat J}^T((X_{\hat J}(m)^TX_{\hat J}(m))^{-1}X_{i,\hat J}(m)$. To show the law convergence in probability to standard normal, we have to show
\begin{align*}
    \mathcal{L}\lrp{\sum_{i=1}^m \hat\alpha_i\hat\epsilon_i^*\mid Y} \xrightarrow[]{Pr}\mathcal{N}(0,1). 
\end{align*}

It is clear to see
\begin{align}
    \E{\hat\alpha_i\hat\epsilon_i^*\mid Y} = 0\;a.s. 
\end{align}

Consider $Var(\sum_{i=1}^m\hat\alpha_i\hat\epsilon_i^*\mid Y)$, we see
\begin{align}
    Var\lrp{\sum_{i=1}^m\hat\alpha_i\hat\epsilon_i^*\mid Y} &= \sum_{i=1}^m\hat{s}^{-1}a_{\hat J}^T(X_{\hat J}(m)^TX_{\hat J}(m))^{-1}X_{i,\hat J}(m)X_{i,\hat J}(m)^T(X_{\hat J}(m)^TX_{\hat J}(m))^{-1}a_{\hat J}\cdot Var\lrp{\hat\epsilon_i^*\mid Y} \\
    &= \sum_{i=1}^m\hat{s}^{-1}a_{\hat J}^T(X_{\hat J}(m)^TX_{\hat J}(m))^{-1}X_{i,\hat J}(m)X_{i,\hat J}(m)^T(X_{\hat J}(m)^TX_{\hat J}(m))^{-1}a_{\hat J}\hat\sigma^2 \\
    &= \hat{s}^{-1}\hat\sigma^2a_{\hat J}^T(X_{\hat J}(m)^TX_{\hat J}(m))^{-1}a_{\hat J} \\
    &= \lrp{a^T_{\hat J}(X_{\hat J}(m)^TX_{\hat J}(m))^{-1}a_{\hat J}}^{-1}a_{\hat J}^T(X_{\hat J}(m)^TX_{\hat J}(m))^{-1}a_{\hat J} =1
\end{align}

The $Var\lrp{\hat\epsilon_i^*\mid Y}=\hat\sigma^2$ because the bootstrap resampling of residuals from $\hat\beta$ will have the variance of $\hat\beta$, which has limiting distribution $\hat{s}^{-1/2}a^T(\hat\beta-\beta)\xrightarrow[]{d}\mathcal{N}(0, 1)$. 

Therefore, we have $\sum_{i=1}^m Var\lrp{\sum_{i=1}^m\hat\alpha_i\hat\epsilon_i^*\mid Y} \xrightarrow[]{Pr} 1$. 

Up to now, we have shown $\E{\hat\alpha_i\hat\epsilon_i^*\mid Y}$ and $Var\lrp{\hat\alpha_i\hat\epsilon_i^*\mid Y}$ converges in probability to $0$ and $1$ respectively. Therefore, to apply the Lindebery-Feller central limit theorem, we only have to verify the Lindeberg condition to show the law converges to standard normal distribution in probability. 

Notice first from the assumption, $\max_{1\leq i\leq m} |\hat\alpha_i|=o_P(1)$. We could see this by
\begin{align}
    \max_{1\leq i\leq m} \hat\alpha_i^2 &= \max_{1\leq i\leq m}\left\{\frac{a_{\hat J}^T(X_{\hat J}(m)^TX_{\hat J}(m))^{-1}X_{i,\hat J}(m)X_{i,\hat J}(m)^T(X_{\hat J}(m)^TX_{\hat J}(m))^{-1}a_{\hat J}}{\hat\sigma^2 a_{\hat J}^T\lrp{X_{\hat J}^T X_{\hat J}}^{-1}a_{\hat J}}\right\} \\
    &\leq \frac{\max_{1\leq i\leq m} \lrn{X_{i,\hat{J}}}^2}{n\hat\sigma^2\rho_1} = O_P\lrp{\frac{p}{n\rho_1}}=o_P(1).
\end{align}

Therefore, we see that 
\begin{align}
    \max_{1\leq i \leq m} \E{|\hat\epsilon^*|^2\ind{|\hat\alpha_i\hat\epsilon^*|\geq \delta}\mid Y} &\leq \E{\hat\epsilon^{*^2} \ind{|\hat\epsilon^*|\max_{1\leq i \leq m} \hat\alpha_i \geq \delta}\mid Y} \\
    &= \E{\hat\epsilon^{*^2} o_P(1)\mid Y} \\
    &= o_P(1).
\end{align}

Compute $\sum_{i=1}^m \hat\alpha_i^2$ under event $\mathcal{E}$, we have
\begin{align}
    \sum_{i=1}^m \hat\alpha_i^2 &= \sum_{i=1}^m\hat{s}^{-1}a_{\hat J}^T(X_{\hat J}(m)^TX_{\hat J}(m))^{-1}X_{i,\hat J}(m)X_{i,\hat J}(m)^T(X_{\hat J}(m)^TX_{\hat J}(m))^{-1}a_{\hat J} \\
    &= \frac{1}{\sigma^2}
\end{align}

Therefore, we know $\sum_{i=1}^m \hat\alpha_i^2\to\frac{1}{\sigma^2}=O(1)$. 

Summarizing the results above, we have the Lindeberg condition that
\begin{align}
    \sum_{i=1}^m \E{|\hat\alpha_i\hat\epsilon^*_i|^2\ind{|\hat\alpha_i\hat\epsilon^*_i|\geq\delta}\mid Y} \xrightarrow[]{Pr} 0\;\;for\;all\;\delta>0.
\end{align}

From  $\E{\hat\alpha_i\hat\epsilon_i^*\mid Y} \xrightarrow[]{Pr} 0$ and $Var\lrp{\hat\alpha_i\hat\epsilon_i^*\mid Y}\xrightarrow[]{Pr} 1$ and the verified Lindeberg condition, we conclude that $\mathcal{L}\lrp{\sum_{i=1}^m \hat\alpha_i\hat\epsilon_i^*\mid Y}\xrightarrow[]{Pr}\mathcal{N}(0,1)$. 



\end{proof}
\section{Proof Theorem \ref{thm:w2_convergence}}
\label{app:proof_w2_convergence}
\begin{proof}
From Thm.~2 in~\citep{ishwaran2005spike}, since our setting implies $\lambda_n/n\to 0$, we see $\Tilde{\beta}_{SSL}=\bar\theta$ asymptotically. 

Thus, from Lemma 2.2, we have
\begin{align}
    \mathcal{L}(\tilde\beta_{SSL}-\beta) \xrightarrow[]{d}\mathcal{N}(0, \bar\sigma^2\lrp{X^TX}^{-1}).
\end{align}

From Thm.~\ref{lem:uq_BIP}, we see from the pivot of Bootstrap replicates
\begin{align}
    \mathcal{L}(\hat\beta^*-\hat\beta) \xrightarrow[]{d}\mathcal{N}\lrp{0, \hat\sigma^{*^2}_m\lrp{X_{J_\beta}^TX_{J_\beta}}^{-1}}
\end{align}

Further, from Thm.~2.1~\citep{giurcanu2016thresholding}, we see
\begin{align}
    \mathcal{L}(\hat\beta-\beta) \xrightarrow[]{d}\mathcal{N}\lrp{0, \hat\sigma^2\lrp{X_{J_\beta}^TX_{J_\beta}}^{-1}}
\end{align}

To show the convergence in Wasserstein distance, we first recall the definition
\begin{align}
    \mathcal{W}_2^2\lrp{\mu,\nu} = \inf_{\pi\in\Pi\lrp{\mu,\nu}} \Ep{(x,y)\sim\pi}{\lrn{x-y}_2^2}.
\end{align}

The Wasserstein distance considers the infimum of all push-forward mappings that assigns $\mu$ to $\nu$. In our case, we consider a trivial push-forward mappings that assigns $\mu$ to $\nu$ uniformly over the entire region. In this way, it just simply considers the difference of two laws. This trivial push-forward mapping is certainly an upper bound of the Wasserstein distance from the definition. Therefore, we see the following: 

We then see the following
\begin{align}
    \mathcal{W}_2^2(\mathcal{L}(\hat\beta^*), \mathcal{L}(\tilde\beta_{SSL})) &\leq \mathcal{W}_2^2(\mathcal{L}(\hat\beta^*), \mathcal{L}(\hat\beta)) + \mathcal{W}_2^2(\mathcal{L}(\hat\beta), \delta_\beta) + \mathcal{W}_2^2(\mathcal{L}(\tilde\beta_{SSL}), \delta_\beta) \\
    &\leq  \E{\lrn{\hat\beta^* - \hat\beta}^2}  + \E{\lrn{\hat\beta - \beta}^2} + \E{\lrn{\tilde\beta_{SSL}-\beta}^2}. 
\end{align}

Then, since all the pivot distributions are asymptotically normal. We could apply the multivariate sub-Gaussian tail results that
\begin{align}
     \mathcal{W}_2^2(\mathcal{L}(\hat\beta^*), \mathcal{L}(\tilde\beta_{SSL})) &\leq   \E{\lrn{\hat\beta^* - \hat\beta}^2}  + \E{\lrn{\hat\beta - \beta}^2} + \E{\lrn{\tilde\beta_{SSL}-\beta}^2}\\
     [By\;Lem.~\ref{lem:subGaussian_mtx}]&\leq 4p\lrp{\lrn{\hat\sigma^{*^2}_m\lrp{X_{J_\beta}^TX_{J_\beta}}^{-1}}_{op}+\lrn{\hat\sigma^{*^2}_m\lrp{X_{J_\beta}^TX_{J_\beta}}^{-1}}_{op}+\lrn{\bar\sigma^2\lrp{X^TX}^{-1}}_{op}}\\
    &\leq 4\frac{p}{n\rho_1}\lrp{\hat\sigma^{*^2}+\hat\sigma^{*^2}_m+\bar\sigma^2}\\
    &= 4\frac{p}{n\rho_1}\lrp{3\sigma^{*^2}+o_P(1)}\\
    [from\;(C.2)] &= 0
\end{align}

From the sub-Gaussian results, we see the Wasserstein distance is upper bounded by $\frac{p}{n\rho_1}$ times a constant factor. From the asymptotic assumption (C.2), we could conclude the convergence in the Wasserstein distance. 

\end{proof}
\section{Technical Lemmas}

\begin{lemma}
\label{lem:subGaussian_mtx}
$X\in\real$ is a sub-Gaussian random vector with parameter $\lrn{\Sigma}_{op}$ if $X\sim\mathcal{N}(0,\Sigma)$. 
\end{lemma}

\section{Implementation of Out-of-bag weighted bagging inclusion probability selection}
\label{app:oob_implementation}
\begin{algorithm*}
    \caption{{Bagging inclusion probability thresholding least-square estimation}}
    \label{alg:bag_tlse_oob}
    \begin{algorithmic}[1]
    \INPUT{covariate $X$, target $y$, thresholding constant $\gamma$, cv proportion $c$, inclusion probability threshold $p$.}
    \OUTPUT{a BIP estimate of active subset $\tilde J$}
    \Function{BIPVariableSelection}{$X, y$}
        \State $MSEs=List()$;   \Comment{create an empty list}
        \State $IncludeList=List()$;   \Comment{create an empty list}
        \For{i in $0, 1, \cdots, n-1$:}
            \State $X_{train}, X_{test}, y_{train}, y_{test} = CrossValidate(X, y, c)$;
            \State $\beta_{BS} = LeastSquares(X_{train}, y_{train})$; \Comment{compute least-square estimate given train data}
            \State $threshold = \sigma\sqrt{\frac{\gamma}{\text{diag}(X_{train}^T X_{train})}}$;  \Comment{threshold array for $\beta_{BS}$ from train data}
            \State $include = (\beta_{BS} > threshold)$;  \Comment{decide whether to include a variable or not from threshold}
            \State $MSE = \frac{1}{cn}\lrn{y_{test}-X_{test}\beta_{BS}}_2^2$;  \Comment{compute mean-squared error}
            \State $IncludeList.append(include)$;
            \State $MSEs.append(MSE)$;
        \EndFor
        \State $Weights = \frac{exp(-MSEs)}{\sum_{i=1}^n(exp(-MSEs[i]))}$; \Comment{normalize the MSE list after exponential. }
        \State $IncludeWeighted = \sum_{i=1}^n Weights[i]\times IncludeList[i]$;
        \State $Selection = IncludeWeighted > p$; \Comment{select indices based on inclusion probability threshold $p$}
        \Return $Selection$
    \EndFunction
    \end{algorithmic}
\end{algorithm*}

\section{Additional plottings}
\begin{figure}
    \centering
    \begin{overpic}[width=0.95\textwidth]{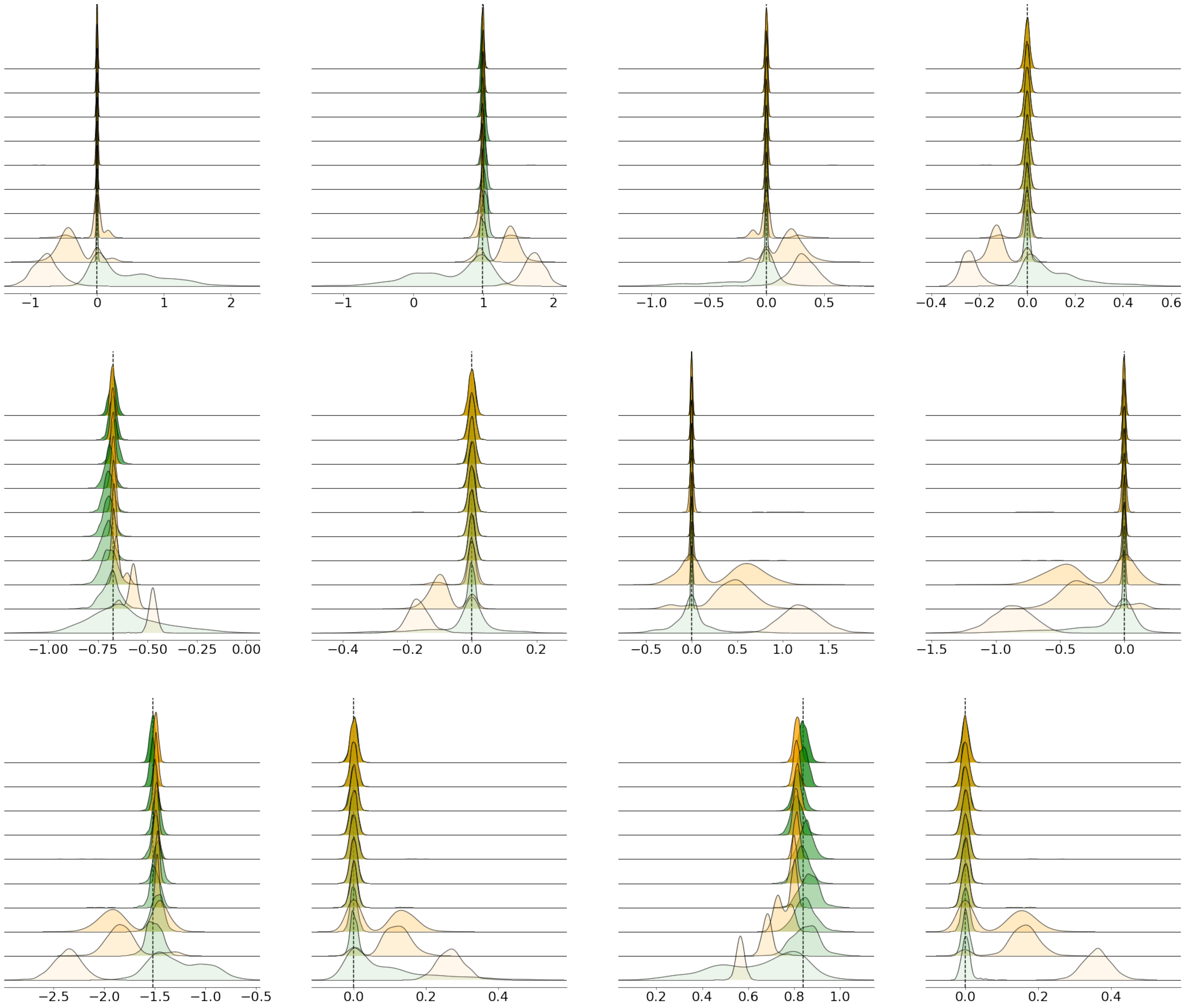}
    \put(7,-1.5){\footnotesize Coefficient}
    \put(33,-1.5){\footnotesize Coefficient}
    \put(59,-1.5){\footnotesize Coefficient}
    \put(84,-1.5){\footnotesize Coefficient}
    \put(-1.2,4.1){\tiny $75$}
    \put(-2,8.1){\tiny $125$}
    \put(-2,12.1){\tiny $175$}
    \put(-2,16.2){\tiny $225$}
    \put(-2,20.3){\tiny $275$}
    \put(-1.2,33.3){\tiny $75$}
    \put(-2,37.3){\tiny $125$}
    \put(-2,41.3){\tiny $175$}
    \put(-2,45.4){\tiny $225$}
    \put(-2,49.5){\tiny $275$}
    \put(-1.2,62.5){\tiny $75$}
    \put(-2,66.5){\tiny $125$}
    \put(-2,70.5){\tiny $175$}
    \put(-2,74.6){\tiny $225$}
    \put(-2,78.7){\tiny $275$}
    \end{overpic}
    \vspace{0.4mm}
    \caption{Extended plottings of Lotka–Volterra experiment with E-SINDy and Bayesian SINDy. }
    \label{fig:convergence_plot_all}
\end{figure}
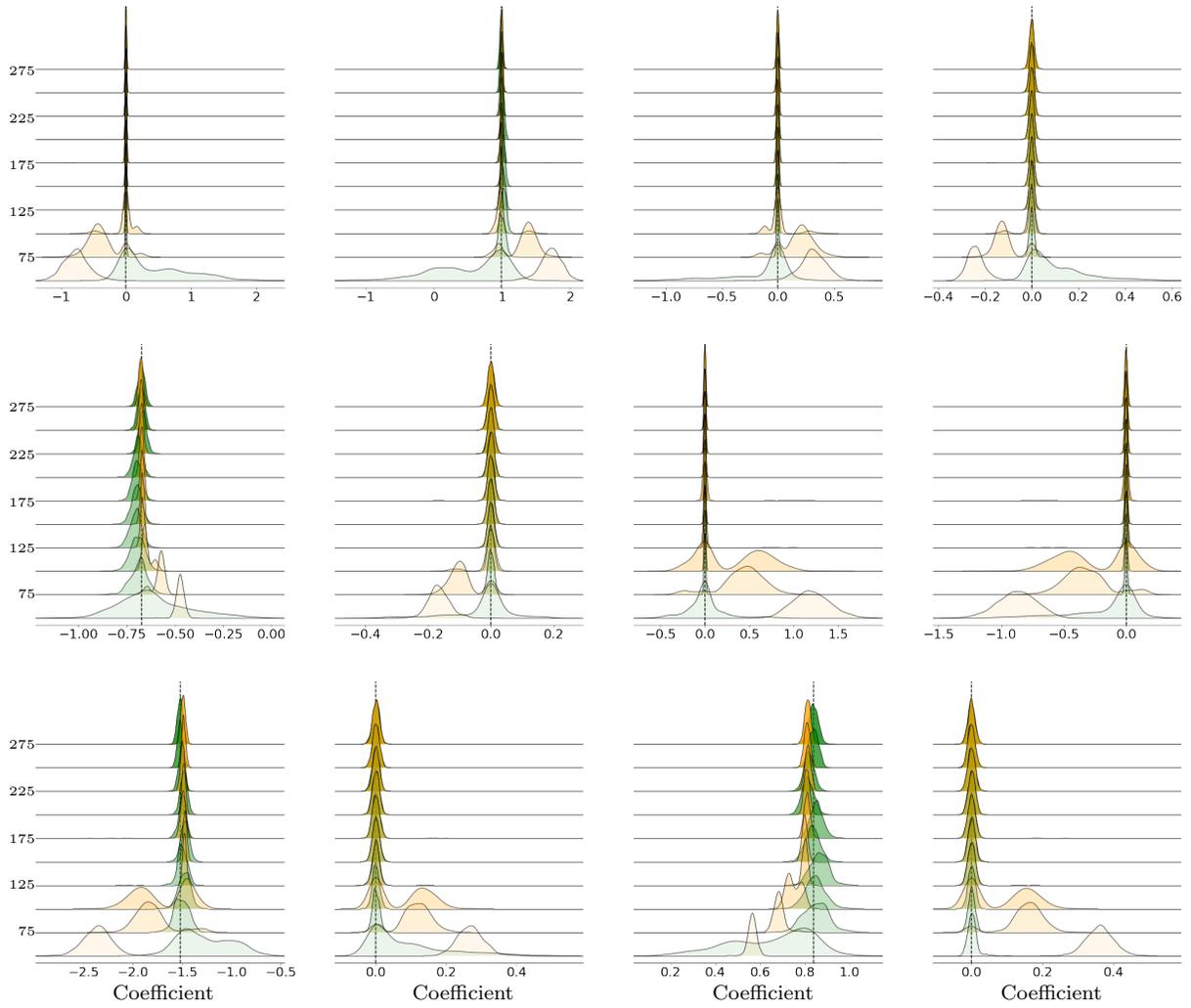
\end{document}